\pdfoutput=1
\documentclass[11pt,letterpaper]{article}

\usepackage[T1]{fontenc}
\usepackage[utf8]{inputenc}
\usepackage{mathpazo}
\usepackage[left=0.92in,right=0.92in,top=0.92in,bottom=0.95in,headheight=22pt]{geometry}
\usepackage{microtype}
\usepackage{amsmath,amssymb,amsfonts}
\usepackage{booktabs}
\usepackage{multirow}
\usepackage{tabularx}
\usepackage{makecell}
\usepackage{array}
\usepackage{placeins}
\usepackage{xurl}
\usepackage{graphicx}
\usepackage{xcolor}
\usepackage{colortbl}
\usepackage{algorithm}
\usepackage{algpseudocode}
\usepackage[numbers,sort&compress]{natbib}
\usepackage{titlesec}
\usepackage{fancyhdr}
\usepackage{caption}

\definecolor{arxivPrimary}{HTML}{2B4F73}
\definecolor{arxivAccent}{HTML}{875961}
\definecolor{arxivMuted}{HTML}{51575E}
\definecolor{biarbiterRow}{HTML}{EAF1F8}
\definecolor{abstractBg}{HTML}{F4F7FA}
\definecolor{abstractRule}{HTML}{D0D5DC}

\usepackage[colorlinks=true,linkcolor=arxivPrimary,citecolor=arxivPrimary,urlcolor=arxivPrimary,pdftitle={BiArbiter: Anchored One-Pass Bidirectional Multimodal Scoring for Unified Routing, Verification, and Decision-Making},pdfauthor={Haoyue Peng, Yuan Liu, Fanliang Hu, Lei Huang, Wenli Cai},pdfsubject={Computation and Language (cs.CL); Artificial Intelligence (cs.AI); Computer Vision and Pattern Recognition (cs.CV)},pdfkeywords={Knowledge-based decision systems, Bidirectional attention, Multimodal candidate scoring, Selective routing, Agent verification, Probability calibration}]{hyperref}

\titleformat{\section}{\large\bfseries\color{arxivPrimary}}{\thesection}{0.6em}{}
\titleformat{\subsection}{\normalsize\bfseries\color{black!88}}{\thesubsection}{0.55em}{}
\titlespacing*{\section}{0pt}{14pt plus 3pt minus 2pt}{5pt plus 1pt}
\titlespacing*{\subsection}{0pt}{10pt plus 2pt minus 1pt}{3pt plus 1pt}
\renewcommand{\arraystretch}{1.10}
\newcommand{\tableformat}{\linespread{1}\small}
\begin{document}

\thispagestyle{plain}
\noindent{\color{arxivPrimary}\rule{\linewidth}{1.6pt}}
\vspace{8pt}
\begin{center}
    {\Large\bfseries\color{black!92} BiArbiter: Anchored One-Pass Bidirectional Multimodal Scoring for Unified Routing, Verification, and Decision-Making\par}
    \vspace{11pt}
    {\normalsize
        \textbf{Haoyue Peng},\;\; \textbf{Yuan Liu},\;\; \textbf{Fanliang Hu},\;\; \textbf{Lei Huang},\;\; \textbf{Wenli Cai}$^{*}$
    \par}
    \vspace{5pt}
    {\small\color{arxivMuted}
        Medical Artificial Intelligence Laboratory, Chongqing University Cancer Hospital,\\
        Chongqing University, Chongqing 400030, China\\[2pt]
        $^*$Corresponding author: \texttt{wenli.cai@cqu.edu.cn}.
    \par}
    \vspace{4pt}
    {\footnotesize\color{arxivMuted} Preprint \;\textbar\; November 30, 2026}
\end{center}
\vspace{4pt}
\noindent{\color{abstractRule}\rule{\linewidth}{0.7pt}}
\vspace{8pt}

\noindent\fcolorbox{abstractRule}{abstractBg}{%
\begin{minipage}{\dimexpr\linewidth-2\fboxsep-2\fboxrule\relax}
\vspace{4pt}
\noindent{\small\bfseries\color{arxivPrimary}Abstract}\\[3pt]
{\small
Knowledge-based pipelines rely on four coupled decision primitives---checker routing, candidate ranking, trajectory verification, and multimodal choice. Current systems either deploy fragmented specialists or invoke generative chain-of-thought judges with multi-second latency and poor score calibration, while naive fine-tuning causes parameter drift and erodes unseen-domain transfer. We propose \textbf{BiArbiter-4B} on hybrid-attention \texttt{H2O-Lightning-4B}, unmasking its eight full-attention layers for bidirectional one-pass scalar readout from two terminal vocabulary rows without autoregressive decoding. Four-stage zero-overlap low-rank adaptation with asymmetric negative gating and $L_2$ parameter anchoring unifies all four primitives in one frozen 4B checkpoint, cutting squared weight drift by 30.6\%. At 1,024 tokens, scoring takes 132.6\,ms (40.4$\times$--80.1$\times$ faster than same-backbone chain-of-thought judging) with native expected calibration error of 0.030--0.043 on routing, research verification, and visual choice. Across 11 benchmarks, it beats the frozen bidirectional base on RewardBench v1 (89.09\%), RuVerBench AgenticCoding (65.77\%), and InFoBench routing (0.8584 Macro-F1; $-75.2\%$ calls), matches base on RewardBench 2, reaches 88.45\% on unseen MMBench TEST, resolves 74.47\% of held-out \texttt{SWE-agent} issues ($-48.4\%$ rollouts), and transfers zero-shot to unseen visual domains with quantified boundaries.
}
\vspace{4pt}
\end{minipage}%
}
\vspace{6pt}

\noindent{\footnotesize\textbf{\color{arxivPrimary}Keywords:} Knowledge-based decision systems $\cdot$ Bidirectional attention $\cdot$ Multimodal candidate scoring $\cdot$ Selective routing $\cdot$ Agent verification $\cdot$ Probability calibration}
\vspace{6pt}

\section{Introduction}
\label{sec:intro}

Knowledge-based systems increasingly couple large language models with retrieval engines, deterministic verifiers, coding agents, and multimodal modules~\cite{yang2025llmkbs}. At each control point in such pipelines, the system must score a candidate artifact against a task specification to execute one of four discrete decisions: dispatching specialized verification checkers (\emph{routing})~\cite{qin2024infobench}, selecting the strongest response or retrieved evidence (\emph{ranking})~\cite{lambert2025rewardbench,hu2025mragbench}, deciding whether an agent trajectory satisfies a checklist criterion or can terminate early (\emph{gating})~\cite{yang2024sweagent,peng2026ruverbench_paper,peng2026ruverbench}, or choosing an answer grounded in visual inputs (\emph{multimodal choice})~\cite{liu2024mmbench}.

\begin{figure}[tbp]
    \centering
    \includegraphics[width=0.98\linewidth]{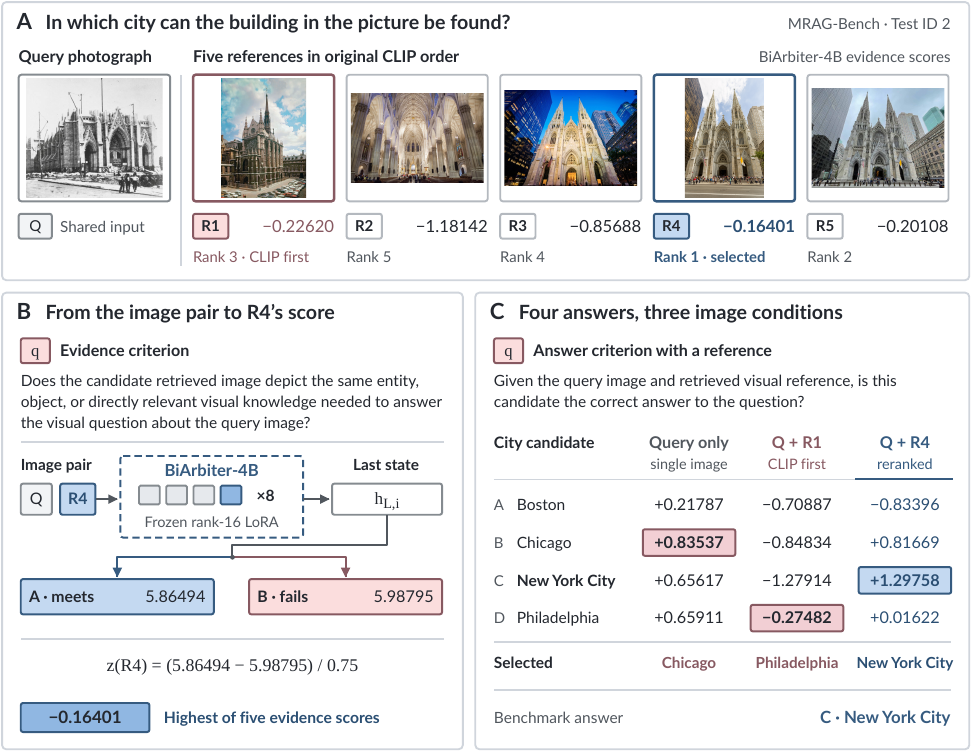}
    \caption{Visual evidence reranking on MRAG-Bench test item 2~\cite{hu2025mragbench}. (A) R4 ranks first among five references with negative evidence scores. (B) Two vocabulary logits from the same terminal state give R4's score. (C) The selected city changes from Philadelphia with R1 to New York City with R4. Table~\ref{tab:case_trace} lists the 17 replay scores. This is the smallest ID among 15 items for which both evidence and answer were corrected. Across all 1,353 items, reranking changed evidence hit@1 from 32.59\% to 36.44\% and answer accuracy from 52.77\% to 51.88\%.}
    \label{fig:case}
\end{figure}

In practice, these primitives frequently interact on the same input. Figure~\ref{fig:case} illustrates a visual retrieval-augmented question where first \emph{ranking} candidate reference photographs directly determines the downstream \emph{multimodal choice}---changing the top-ranked reference flips the predicted city to the ground-truth answer using one shared frozen scorer.

Existing pipelines fragment these decisions across incompatible paradigms. Dedicated scalar reward models perform well on text preference ranking~\cite{wang2024armorm,wang2025helpsteer2pref} but cannot ingest images, execute multi-label checker routing, or provide calibrated zero-threshold gating. Conversely, generative chain-of-thought (CoT) judges~\cite{zheng2023judging} generalize through natural-language prompts, yet generating 128--256 reasoning tokens per candidate incurs multi-second latency that precludes interactive routing, multi-candidate reranking, and per-turn agent monitoring. Meanwhile, directly fine-tuning compact 4B--8B models for specific control tasks induces parameter drift and catastrophic forgetting~\cite{kirkpatrick2017ewc}, eroding zero-shot transfer to unseen domains.

This paper investigates a unified alternative: \emph{can a single 4B-parameter multimodal scorer serve all four decision primitives in one forward pass, output natively calibrated confidence scores for selective abstention and early stopping, and preserve zero-train transferability to unseen application domains?}

We introduce \textbf{BiArbiter-4B}, built on the 32-layer hybrid-attention \texttt{H2O-Lightning-4B} backbone~\cite{h2o2026lightning}. Two architectural choices govern its accuracy and speed. First, rather than enforcing causal masking---which prevents earlier candidate and image tokens from reading a trailing evaluation criterion---we unmask all eight full-attention layers to bidirectional attention (\texttt{is\_causal=False}) while retaining the 24 linear-attention layers. Second, instead of autoregressive decoding, \textbf{BiArbiter-4B} extracts a scalar logit at the terminal prompt token via two 32-bit floating-point (FP32) vocabulary rows, $z = (h_L^\top w_{\mathrm{A}} - h_L^\top w_{\mathrm{B}})/\tau_0$ ($\tau_0 = 0.75$). This one-pass readout scores a 1,024-token candidate in 132.6\,ms (89.0\,ms for a 512\,px vision-language input), running 40.4$\times$--80.1$\times$ faster than the same 4B backbone executing 128--256-token CoT judging.

Adapting this bidirectional backbone into a calibrated multi-primitive arbiter requires overcoming two frozen-model failure modes: severe threshold miscalibration---seen in empty-dispatch collapse on InFoBench routing (Macro-F1 0.027--0.174) and over-positive bias on RuVerBench AgenticCoding (only 28.13\% macro negative recall)---and cross-domain forgetting under sequential updates. We address both through an \emph{anchored multi-stage calibration} pipeline under strict SHA-256 prompt, instance, and image isolation. Starting from a merged Round-1 bidirectional low-rank adaptation (LoRA)~\cite{hu2022lora} initialization, we train a single 7.67M-parameter LoRA adapter across four zero-overlap stages combining conflict-balanced checker rosters, hard negatives, adversarial instruction-following/refusal pairs, and real software-engineering trajectories with asymmetric negative gate weighting, within-set pairwise margins, and an explicit $L_2$ parameter anchor ($\lambda_{\mathrm{anc}}\|\theta - \theta_{\mathrm{ref}}\|_2^2$).

We evaluate this single frozen checkpoint across 11 benchmarks spanning public leaderboards, execution-verified closed loops, and three held-out visual, document, and industrial domains. Our main contributions are:

\begin{enumerate}
    \item \textbf{Unified One-Pass Bidirectional Scoring.} We formulate checker routing, candidate ranking, verification gating, and multimodal choice under one multimodal scoring interface. Matched causal-versus-bidirectional ablations confirm consistent accuracy gains both zero-shot and after fine-tuning while eliminating autoregressive judging latency.
    \item \textbf{Anchored Multi-Stage Calibration and Mechanism Ablations.} We pair asymmetric negative gate supervision with $L_2$ parameter anchoring across four zero-overlap stages. Single-factor ablations verify that anchoring cuts squared LoRA drift by 30.6\% while preventing RewardBench degradation, asymmetric gating restores negative trajectory recall, and blind-image testing confirms genuine visual grounding ($+55.93$ percentage points on unseen MMBench).
    \item \textbf{Native Calibration and Closed-Loop Downstream Gains.} Without post-hoc temperature rescaling, \textbf{BiArbiter-4B} attains low native Expected Calibration Error (ECE, 0.030--0.043) across routing, deep research verification, and unseen visual choice. In live execution loops, it cuts InFoBench checker calls by 75.2\% at 0.8584 Macro-F1 and raises disjoint \texttt{SWE-agent} unit-test resolution from 43.62\% (Pass@1) to 75.53\% (Best-of-8), retaining 74.47\% with 48.4\% fewer rollouts via early stopping.
    \item \textbf{Zero-Train Cross-Domain Transfer and Boundary Analysis.} Using identical frozen weights on unseen application domains, \textbf{BiArbiter-4B} improves MRAG-Bench retrieval, routes multipage MP-DocVQA evidence pages (74.14\% vs.\ 28.67\% first-page), and detects industrial MVTec-AD anomalies (0.8678 macro area under the ROC curve, AUROC). Equally, we quantify capability boundaries, including a 0.3--1.9 percentage-point visual prior drift relative to the unfine-tuned base, RewardBench 2 ties-versus-margin sensitivity, and the rubric gap on RubricBench.
\end{enumerate}

\section{Related Work}
\label{sec:related}

\subsection{Reward Models and LLM-as-a-Judge Evaluators}
Learned preference models anchor reinforcement learning from human feedback~\cite{ouyang2022instructgpt} and test-time verification~\cite{lightman2024lets}. Classical scalar reward models adopt Bradley--Terry pairwise scoring~\cite{bradley1952rank}, whereas implicit objectives such as DPO~\cite{rafailov2023dpo} couple reward estimation to policy likelihoods, and mutual-supervision frameworks further improve reward robustness~\cite{peng2026mutual}. Recent benchmarks expose complementary evaluator failure modes: RewardBench v1~\cite{lambert2025rewardbench} probes chat, adversarial instruction following~\cite{zeng2024llmbar}, safety refusals~\cite{rottger2024xstest}, and reasoning; RewardBench 2~\cite{malik2026rewardbench2} extends evaluation to multi-candidate Factuality, Precise IF, Math, Focus, and Ties; RuVerBench~\cite{peng2026ruverbench_paper,peng2026ruverbench} evaluates long-horizon rubric verification over deep-research reports and agentic coding trajectories; and RubricBench~\cite{zhou2026rubricbench} isolates surface-bias distractors when explicit human rubrics are withheld.

Specialized text reward models---including ArmoRM-8B~\cite{wang2024armorm}, HelpSteer2-Preference models~\cite{wang2024helpsteer2,wang2025helpsteer2pref}, debiased preference datasets~\cite{park2024offsetbias}, safety moderation suites~\cite{han2024wildguard}, and post-fine-tuning safety alignment methods~\cite{jiang2025postft}---excel on text ranking and safety, yet remain predominantly unimodal 8B--70B specialists. Conversely, generative LLM-as-a-Judge evaluators~\cite{zheng2023judging} generalize across tasks via natural-language prompts, but autoregressive decoding incurs multi-second latency per candidate that bottlenecks multi-checker routing, retrieval reranking, and per-step agent verification. \textbf{BiArbiter-4B} bridges both paradigms by pairing natural-language criterion conditioning with one-pass multimodal scalar readout (Figure~\ref{fig:comparison} contrasts causal generation against one-pass scoring under matched hardware and backbone controls).

\subsection{Bidirectional Attention in Decoder Backbones and Multimodal Decisions}
Causal masking in standard decoder-only transformers prevents earlier candidate or image tokens from attending to trailing verification criteria. Removing this mask in decoder encoders such as LLM2Vec~\cite{behnamghader2024llm2vec} substantially strengthens retrieval and classification representations, and \texttt{H2O-Lightning-4B}~\cite{h2o2026lightning} builds a typed multimodal decision backbone atop \texttt{Qwen3.5-4B} supporting binary and multiple-choice readouts. Recent multimodal knowledge pipelines likewise target retrieval-augmented visual QA via collaborative parametric calibration~\cite{deng2026collabvqa} or training-free zero-shot anomaly detection over unverified evidence~\cite{yun2026queco}. Whereas MMBench~\cite{liu2024mmbench}, MRAG-Bench~\cite{hu2025mragbench}, MP-DocVQA~\cite{tito2023mpdocvqa}, and MVTec-AD~\cite{bergmann2019mvtec} are typically handled by separate task-specific models, we isolate the causal-versus-bidirectional effect on a hybrid linear/full-attention 4B backbone and test whether one adapted checkpoint can unify text routing, agent verification, visual choice, and zero-train visual transfer.

\subsection{Selective Routing, Calibration, and Anchored Adaptation}
Cost-aware cascades such as FrugalGPT~\cite{chen2024frugalgpt} and RouteLLM~\cite{ong2025routellm} cut inference cost by dispatching queries across model tiers, while failure-mode-aware uncertainty routing ties intervention directly to anticipated errors~\cite{yin2026uncertaintyrouting}. Safe routing and abstention require well-calibrated probabilities~\cite{guo2017calibration} and low selective risk under coverage budgets~\cite{geifman2017selective}. Meanwhile, sequential low-rank adaptation (LoRA)~\cite{hu2022lora} risks catastrophic forgetting of prior skills~\cite{kirkpatrick2017ewc}, prompting continual-learning systems to maintain multiple task-specific adapters~\cite{jin2025multilora}. Instead of multiplying adapters, \textbf{BiArbiter-4B} maintains one frozen LoRA checkpoint by pairing an explicit $L_2$ weight anchor to the prior stage with balanced replay and asymmetric negative gate calibration, preserving native logit calibration (ECE 0.030--0.043 across routing, deep-research verification, and unseen visual choice) for direct escalation and early stopping.

\section{Methodology: The BiArbiter-4B Framework}
\label{sec:method}

\begin{figure}[tbp]
    \centering
    \includegraphics[width=0.98\linewidth]{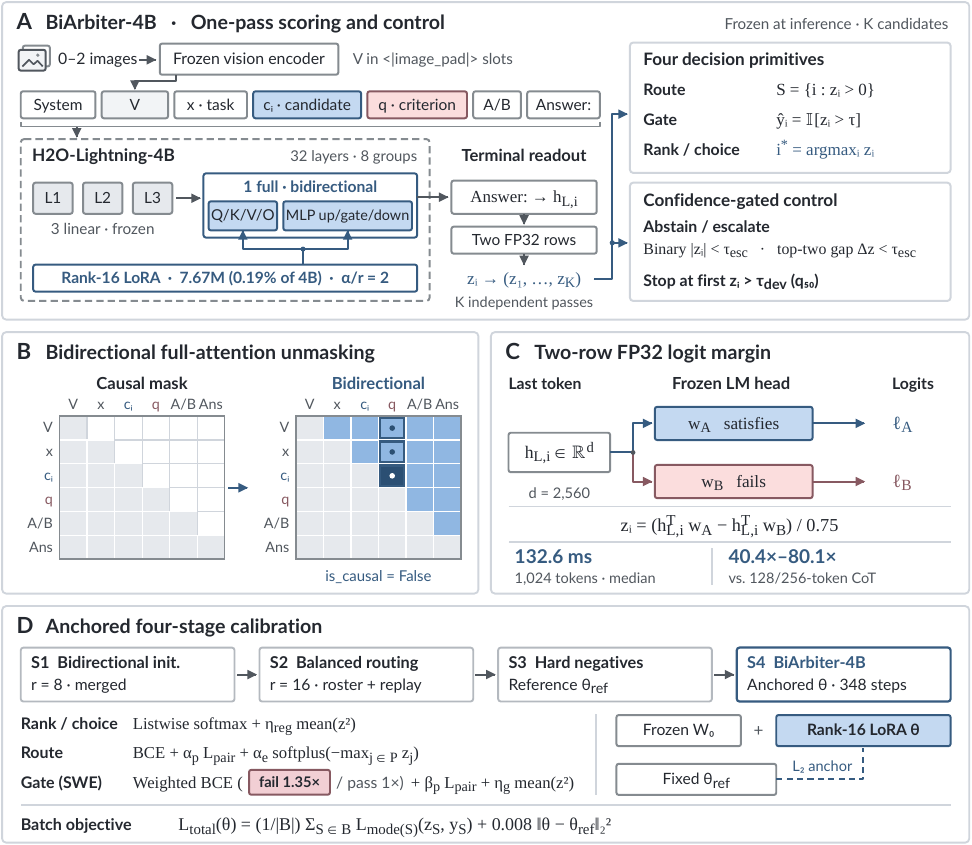}
    \caption{BiArbiter-4B scores each candidate in one forward pass. (A) Packed multimodal prompts pass through eight groups of three frozen linear-attention layers and one unmasked full-attention layer (with rank-16 LoRA on attention and full-block MLPs, 7.67M trainable parameters); the resulting scalar margins drive four unified decision primitives plus selective escalation and sequential early stopping. (B) Unmasking the eight full-attention layers allows visual tokens $V$, task $x$, and candidate $c_i$ to attend to the trailing criterion $q$. (C) Two frozen vocabulary rows project the terminal state $h_{L,i}$ to the FP32 margin without autoregressive decoding. (D) Four-stage zero-overlap calibration anchors Stage~4 weights $\theta$ to the Stage~3 midpoint $\theta_{\mathrm{ref}}$ while optimizing primitive-aligned objectives.}
    \label{fig:framework}
\end{figure}

Figure~\ref{fig:framework} summarizes the full \textbf{BiArbiter-4B} architecture, spanning one-pass bidirectional scoring, unified four-primitive dispatch and confidence control, two-row FP32 readout, and anchored four-stage calibration.

\subsection{Unified Four-Primitive Decision Formulation}
\label{sec:formulation}

Let a decision instance consist of a task context $x$, an evaluation criterion or question $q$, a candidate pool $\mathcal{C} = \{c_1, \dots, c_K\}$, and an optional visual context $V \in \emptyset \cup \mathcal{I} \cup \mathcal{I}^2$ (supporting text-only, single-image, or dual-image query-plus-evidence pairs). Rather than introducing task-specific classification heads or domain-specific adapters, \textbf{BiArbiter-4B} maps each candidate independently through a single scoring function $f_\theta$:
\begin{equation}
    z_i = f_\theta(x, q, c_i, V) \in \mathbb{R}, \qquad i \in \{1, \dots, K\}.
    \label{eq:score}
\end{equation}
The resulting vector $\mathbf{z} = (z_1, \dots, z_K)$ drives four distinct downstream decision primitives without changing weights or prompt schema:
\begin{itemize}
    \item \textbf{Multi-Label Routing (\texttt{route}):} Each $c_i$ describes an executable downstream checker (e.g., Content, Format, Linguistic, Number, Style). The system dispatches subset $\mathcal{S} = \{i : z_i > 0\}$, incurring compute proportional to $|\mathcal{S}|/K$.
    \item \textbf{Candidate Ranking (\texttt{rank}):} Given $K \ge 2$ competing responses, retrieved evidence images, or document pages, the system selects $i^\star = \arg\max_i z_i$ or sorts candidates in descending order of $z_i$.
    \item \textbf{Verification Gating and Early Stopping (\texttt{gate}):} Given a candidate report, response, or multi-turn agent trajectory and an atomic criterion $q$, the model outputs acceptance $\hat{y}_i = \mathbb{I}[z_i > \tau]$ (default $\tau=0$ for rubric verification, or a dev-frozen threshold $\tau_{\mathrm{dev}}$ for sequential Best-of-$K$ early stopping).
    \item \textbf{Multimodal Choice (\texttt{choice}):} Given a visual question $(x, V)$ and $K$ candidate answers, the system selects $i^\star = \arg\max_i z_i$ conditioned on real pixel tensors.
\end{itemize}

\subsection{One-Pass Bidirectional Hybrid-Attention Readout}
\label{sec:bidir}

The backbone \texttt{H2O-Lightning-4B} (built on \texttt{Qwen3.5-4B}~\cite{h2o2026lightning}) uses a 32-layer hybrid language architecture comprising 24 linear-attention layers interleaved with 8 full self-attention layers (\texttt{Qwen3\_5Attention}), coupled with a native multimodal vision encoder. Each candidate prompt packs the system instruction, optional vision slots (\texttt{<|image\_pad|>}), task context $x$, candidate $c_i$, criterion $q$, binary option descriptions (\texttt{A} for satisfies, \texttt{B} for does not satisfy), and a deterministic assistant prefill terminating at the token sequence \texttt{Answer:}.

Under standard causal masking, representations of the task context, candidate text, and image patches are computed before observing the criterion $q$ placed near the end of the prompt. Because \textbf{BiArbiter-4B} never appends ground-truth supervision tokens to the input sequence and never performs autoregressive token generation, causal masking is unnecessary during scoring. We register a forward pre-hook on the language model that sets \texttt{is\_causal = False} across all 8 full-attention layers and supplies unpadded scaled dot-product attention (SDPA) masks (\texttt{attention\_mask = None} for both attention families on unpadded microbatches). Every candidate and visual patch token can therefore exchange information directly with the criterion $q$ in one pass.

From the final hidden state $h_L \in \mathbb{R}^d$ at the last prompt position, we project only onto the two vocabulary rows corresponding to options \texttt{A} and \texttt{B}, cast to FP32 for numerical stability:
\begin{equation}
    z_i = \frac{h_L^\top w_{\mathrm{A}} - h_L^\top w_{\mathrm{B}}}{\tau_0},
    \label{eq:readout}
\end{equation}
where $\tau_0 = 0.75$ is the backbone's fixed native logit temperature. Avoiding full-vocabulary softmax and autoregressive KV-cache decoding keeps memory usage under 9.7\,GiB even at 8,192 tokens and delivers a 40.4$\times$--80.1$\times$ speedup over generative CoT judging on the same hardware. For inputs exceeding the context budget ($L_{\max}=9{,}500$ during training, $16{,}384$ at evaluation), text-only prompts apply head-and-tail truncation so that both the initial task specification and the terminal criterion plus \texttt{Answer:} readout token are preserved intact.

\subsection{Anchored Multi-Stage Calibration Training}
\label{sec:training}

To resolve frozen-model threshold bias without eroding general multimodal knowledge, we adapt \texttt{H2O-Lightning-4B} through four strictly isolated stages using LoRA~\cite{hu2022lora} on the language model while keeping the vision encoder frozen:
\begin{enumerate}
    \item \textbf{Stage 1 (Round-1 Bidirectional Initialization):} A fresh rank-8 attention LoRA (1.57M parameters, 255 steps) is trained on the clean four-primitive \texttt{strict v3} dataset (2,033 sets across HelpSteer2, InFoBench, MMBench, and VerInstruct) under bidirectional masking and merged into the base weights.
    \item \textbf{Stage 2 (Conflict-Balanced Checker Roster, denoted Step~6 / \texttt{step6}):} Starting from the merged Stage-1 backbone, we attach a rank-16 attention+MLP LoRA (7,667,712 trainable parameters, 0.19\% of 4B) and train on 700 conflict-balanced routing sets enriched with an explicit five-checker taxonomy roster plus 446 non-route replay sets (two routing epochs paired with one replay pass; 1,846 set exposures, 232 updates). Explicit negative labeling on pure Linguistic/Style requirements (\texttt{Content=0}) resolves content over-dispatch and eliminates empty routing.
    \item \textbf{Stage 3 (Zero-Overlap Hard-Negative Multitask, denoted Step~7 / \texttt{step7\_mid}):} We continue training over 3,200 sets (400 updates) combining hard HelpSteer2 preference margins, RubricHub pairs, IFBench/VerInstruct gates, balanced routes, and MMBench replay, verified for zero overlap against all evaluation splits; Stage~4 inherits the 200-update midpoint checkpoint \texttt{step7\_mid} (1,600 sets seen) as $\theta_{\mathrm{ref}}$.
    \item \textbf{Stage 4 (Targeted Adversarial \& Agent-Trajectory Calibration, \textbf{BiArbiter-4B}):} Starting from \texttt{step7\_mid} ($\theta_{\mathrm{ref}}$), we run 348 cosine-scheduled updates (batch size 8 sets, peak $\text{lr}=1.2\times 10^{-5}$, weight decay $0.01$) on 2,781 zero-overlap sets targeting two diagnosed external weaknesses: (a) adversarial instruction following and benign non-refusal from OffsetBias~\cite{park2024offsetbias} (400 sets), Tulu-3 IF preferences (350 sets), and balanced WildGuard benign/harmful refusal pairs~\cite{han2024wildguard} (300 sets); (b) multi-turn coding-agent verification from \texttt{nebius/\allowbreak SWE-agent-\allowbreak trajectories}~\cite{yang2024sweagent} shards 00--03 split strictly by GitHub \texttt{instance\_id} (200 paired resolved-vs-submitted-failed task sets, 340 single-trajectory sets, and 360 deterministic patch/test/tool-discipline process gates); and (c) 831 multi-primitive anchor replay sets (InFoBench, MMBench, IFBench, VerInstruct, HelpSteer2, RubricHub).
\end{enumerate}

Let $\mathbf{z} \in \mathbb{R}^K$ and $\mathbf{y} \in \{0,1\}^K$ denote the predicted logits and binary labels of a candidate set $S$. Each primitive uses a loss aligned with its downstream operator:
\begin{align}
    \mathcal{L}_{\mathrm{rank/choice}}(\mathbf{z}, \mathbf{y}) &= -\sum_{i: y_i=1} \log \frac{e^{z_i}}{\sum_{j=1}^K e^{z_j}}
        + \frac{\eta_{\mathrm{reg}}}{K}\sum_{j=1}^K z_j^2, \label{eq:loss_rank} \\
    \mathcal{L}_{\mathrm{route}}(\mathbf{z}, \mathbf{y}) &= \mathcal{L}_{\mathrm{BCE}}(\mathbf{z}, \mathbf{y})
        + \alpha_{\mathrm{p}} \mathcal{L}_{\mathrm{pair}}(\mathbf{z}, \mathbf{y}) \nonumber \\
        &\quad + \alpha_{\mathrm{e}}\,\mathrm{softplus}\!\left(-\max_{j \in \mathcal{P}} z_j\right), \label{eq:loss_route} \\
    \mathcal{L}_{\mathrm{gate}}(\mathbf{z}, \mathbf{y}) &= \frac{1}{K}\sum_{j=1}^K w(y_j)\,\ell_{\mathrm{BCE}}(z_j, y_j) \nonumber \\
        &\quad + \beta_{\mathrm{p}} \mathcal{L}_{\mathrm{pair}}(\mathbf{z}, \mathbf{y})
        + \frac{\eta_{\mathrm{g}}}{K}\sum_{j=1}^K z_j^2, \label{eq:loss_gate}
\end{align}
where $\mathcal{L}_{\mathrm{pair}}(\mathbf{z}, \mathbf{y}) = \frac{1}{|\mathcal{P}||\mathcal{N}|}\sum_{i\in\mathcal{P}}\sum_{j\in\mathcal{N}}\log(1 + e^{-(z_i - z_j)})$ enforces positive-over-negative separation whenever both positive ($\mathcal{P}$) and negative ($\mathcal{N}$) candidates coexist in $S$, and $\alpha_{\mathrm{e}}\,\mathrm{softplus}(-\max_{j \in \mathcal{P}} z_j)$ ($\alpha_{\mathrm{e}}=0.25$ whenever $\mathcal{P} \neq \emptyset$, zero otherwise) prevents empty dispatch on routing sets. In Eq.~(\ref{eq:loss_rank}), Stage~4 applies $\eta_{\mathrm{reg}}=0.005$ to \texttt{rank} sets and $\eta_{\mathrm{reg}}=0$ to \texttt{choice} sets. For coding-agent gates, asymmetric negative weighting sets $w(1)=1.0$ and $w(0)=w_{\mathrm{neg}}=1.35$ (1.25 for non-SWE gates) with $\beta_{\mathrm{p}}=0.75$ (0.50 for non-SWE) and $\eta_{\mathrm{g}}=0.002$ ($0$ for non-SWE gates) to counter over-positive trajectory bias.

Crucially, at every optimizer step we add an explicit $L_2$ parameter anchor tying active LoRA weights $\theta$ to the prior reference checkpoint $\theta_{\mathrm{ref}}$:
\begin{equation}
    \mathcal{L}_{\mathrm{total}}(\theta) = \frac{1}{|\mathcal{B}|}\sum_{S \in \mathcal{B}} \mathcal{L}_{\mathrm{mode}(S)}(\mathbf{z}_S, \mathbf{y}_S) + \lambda_{\mathrm{anc}} \|\theta - \theta_{\mathrm{ref}}\|_2^2,
    \label{eq:loss_total}
\end{equation}
with $\lambda_{\mathrm{anc}} = 0.008$. Algorithm~\ref{alg:train} summarizes the anchored update loop and unified inference dispatch.

\begin{algorithm}[t]
\caption{Anchored Bidirectional Calibration and Unified Inference}
\label{alg:train}
\begin{algorithmic}[1]
\Require Reference LoRA weights $\theta_{\mathrm{ref}}$ (\texttt{step7\_mid}), zero-overlap training schedule $\{\mathcal{B}_t\}_{t=1}^T$, anchor weight $\lambda_{\mathrm{anc}}=0.008$.
\State Enable bidirectional masking (\texttt{is\_causal=False}) on all 8 full-attention layers.
\State Initialize trainable LoRA parameters $\theta \leftarrow \theta_{\mathrm{ref}}$ (7.67M params).
\For{$t = 1$ to $T$}
    \State Zero gradients; set cosine learning rate $\eta_t$.
    \For{each candidate set $S = (x, q, \{c_i\}_{i=1}^K, V, \mathbf{y}, m) \in \mathcal{B}_t$}
        \State Compute one-pass logits $z_i = (h_{L,i}^\top w_{\mathrm{A}} - h_{L,i}^\top w_{\mathrm{B}})/\tau_0$ for $i=1..K$.
        \State Accumulate $\frac{1}{|\mathcal{B}_t|}\nabla_\theta \mathcal{L}_m(\mathbf{z}, \mathbf{y})$ via Eqs.~(\ref{eq:loss_rank})--(\ref{eq:loss_gate}).
    \EndFor
    \State Add anchor gradient $\nabla_\theta \left(\lambda_{\mathrm{anc}}\|\theta - \theta_{\mathrm{ref}}\|_2^2\right)$; clip $\|\nabla_\theta\|_2 \le 1.0$; step AdamW.
\EndFor
\State \textbf{Freeze} $\theta^\star$ as \textbf{BiArbiter-4B}. At test time, compute $z_i$ in one pass and apply primitive operator (\texttt{route}: $z_i>0$; \texttt{rank}/\texttt{choice}: $\arg\max z_i$; \texttt{gate}: $z_i>\tau_{\mathrm{dev}}$; escalate if $|z_i|<\tau_{\mathrm{esc}}$).
\end{algorithmic}
\end{algorithm}

\subsection{Confidence-Gated Escalation and Sequential Early Stopping}
\label{sec:selective}

Because \textbf{BiArbiter-4B} outputs calibrated scalar margins without domain-specific rescaling, its raw uncertainty signal ($|z_i|$ for binary gating/routing, or the top-1-minus-top-2 margin $\Delta z = z_{(1)} - z_{(2)}$ for ranking and choice) directly supports two system-level control policies:
\begin{enumerate}
    \item \textbf{Selective Escalation:} Decisions with confidence below a development-fitted threshold $\tau_{\mathrm{esc}}$ are abstained and escalated to a frontier model or human reviewer, trading coverage for sharply reduced error rate (quantified by Area Under the Risk-Coverage Curve, AURC~\cite{geifman2017selective}).
    \item \textbf{Sequential Early Stopping for Coding Agents:} During Best-of-$K$ trajectory sampling, trajectories are generated sequentially and scored in one pass after completion. Sampling terminates immediately upon the first candidate satisfying $z_i > \tau_{\mathrm{dev}}$ (fitted strictly on the independent Stage-4 development split); otherwise the highest-scoring trajectory among all $K$ rollouts is submitted.
\end{enumerate}

\section{Experimental Setup}
\label{sec:setup}

\subsection{Benchmarks, Zero-Overlap Audits, and Evidence Boundaries}
\label{sec:benchmarks}

Following strict separation of evidence boundaries, we organize our 11 evaluation suites into five distinct categories. No benchmark test split, leaderboard item, or held-out domain was ever used for training, checkpoint selection, prompt tuning, or threshold fitting.

\begin{enumerate}
    \item \textbf{Internal Locked Multi-Primitive Holdout (\texttt{v3\_holdout} \& \texttt{inc\_holdout}):} 373 locked decision sets (InFoBench routing: 22 tasks / 80 requirements / 400 candidate scores; HelpSteer2 preference: 164 pairs; MMBench visual choice: 120 images; VerInstruct gate: 9 items) plus 47 incremental Linguistic/Style conflict sets.
    \item \textbf{Official External Benchmarks:}
    \begin{itemize}
        \item \emph{RewardBench v1}~\cite{lambert2025rewardbench}: Full 2,985 filtered pairs across 23 subsets (Chat, Chat Hard, Safety, Reasoning; 0 truncations).
        \item \emph{RewardBench 2}~\cite{malik2026rewardbench2}: Full 1,865 prompts / 8,977 candidates across 6 categories (Focus, Safety, Ties, Math, Factuality, Precise IF; 0 truncations), scored locally by the official evaluator.
        \item \emph{RuVerBench}~\cite{peng2026ruverbench,peng2026ruverbench_paper}: Full 494 cases / 2,458 atomic rubric checks, covering DeepResearch (284 reports / 1,615 rubrics) and AgenticCoding (210 \texttt{Claude-Code} trajectories / 843 checks), scored with the official \texttt{compute\_main\_leaderboard.py} macro balanced accuracy (BAcc) script.
        \item \emph{MMBench EN V1.1}~\cite{liu2024mmbench}: Both DEV (1,292 CircularEval groups / 4,876 rows) and official TEST (1,925 groups / 7,299 rows). Because 702 DEV groups and 29 TEST groups share exact image hashes or question IDs with \texttt{strict v3} training, we explicitly report strictly unseen DEV (\textbf{590 groups}, and \textbf{326 groups} outside all v3 train/dev/holdout) and strictly unseen TEST (\textbf{1,896 groups}).
    \end{itemize}
    \item \textbf{Execution-Verified Closed-Loop Downstream Utility:}
    \begin{itemize}
        \item \emph{InFoBench Checker Routing (C1):} Measures actual checker invocation savings versus Always-All (5 checkers/req), Content-Only, Train-Frequency, and matched-budget random baselines.
        \item \emph{SWE-agent Best-of-$K$ and Early Stopping (C2):} Drawn exclusively from newly downloaded \texttt{nebius/\allowbreak SWE-agent-\allowbreak trajectories}~\cite{yang2024sweagent,jimenez2024swebench} shards 04--05, strictly excluding all 1,213 \texttt{instance\_id}s present in shards 00--03 and all 582 Stage-4 train/dev instances (0 shared instances). Yields 268 completely unseen GitHub issues with $K=8$ rollouts each, evaluated by ground-truth unit-test pass rate on both a natural-distribution pool ($n=200$) and a discriminative mixed pool ($n=94$ issues containing both passing and failing rollouts).
    \end{itemize}
    \item \textbf{True Held-Out Zero-Train Cross-Application Domains:} Three visual, document, and industrial benchmarks absent from all four training stages, evaluated with identical frozen weights:
    \begin{itemize}
        \item \emph{MRAG-Bench}~\cite{hu2025mragbench} (ICLR 2025, full 1,353 test items; 731 mixed retrieval cases containing both ground-truth and distractor images): Dual-image visual evidence reranking over 5 CLIP-retrieved candidates (Top-1 Hit, MRR, NDCG@5) and end-to-end retrieval-augmented visual QA.
        \item \emph{MP-DocVQA}~\cite{tito2023mpdocvqa} (5,187 validation questions / 28,906 candidate pages; 3,554 multipage questions): Page-level visual evidence routing (Top-1, MRR vs.\ first-page and random).
        \item \emph{MVTec-AD}~\cite{bergmann2019mvtec} (198 test images across \texttt{bottle} and \texttt{metal\_nut}, 0 image-hash overlap with training): Zero-shot industrial anomaly gating (macro AUROC, BAcc, F1).
    \end{itemize}
    \item \textbf{Held-Out Constraint and Surface-Bias Stress Tests:}
    \begin{itemize}
        \item \emph{IFBench}~\cite{pyatkin2025ifbench} (295 unseen test items after excluding 5 items matching training prompts): Strict and loose verifiable constraint gating (AUROC, BAcc).
        \item \emph{RubricBench}~\cite{zhou2026rubricbench} (1,146 strictly zero-overlap pairs after excluding 1 training-prefix match): Pairwise ranking without expert rubrics across five domains under length, tone, and formatting distractors.
    \end{itemize}
\end{enumerate}

\subsection{Implementation, Baselines, and Statistical Protocol}
\label{sec:impl}

All training and inference runs start from the released \texttt{H2O-Lightning-4B} v1.4.2 checkpoint (Hugging Face commit \texttt{839c5c95}, \texttt{model.safetensors} SHA-256 prefix \texttt{c9b6290d})~\cite{h2o2026lightning} and execute in a pinned container on a single NVIDIA A100-SXM4-80GB MIG \texttt{3g.40gb} slice using BF16 weights and SDPA attention. Training peak memory stays within 13.53\,GiB; inference uses 8.62--9.67\,GiB. Stage 4 (\textbf{BiArbiter-4B}) trains in 4,763\,s (1.32\,h). Zero-overlap SHA-256 verification checks confirm 0 shared prompts, hashes, images, or GitHub instances across all audited evaluation boundaries (audit summary in Table~\ref{tab:app_cross}).

To avoid inflating statistical power by treating correlated items (e.g., multiple rubrics per report or circular permutations of the same image) as independent observations, every significance test aggregates strictly at the independent group level (prompt, case, task, issue, or image group). We compute 10,000-resample paired bootstrap 95\% confidence intervals (2,000 resamples for weighted RewardBench v1 overall and RuVerBench case-cluster balanced accuracy) together with exact two-sided McNemar tests (for binary group outcomes) or two-sided paired sign-flip permutation tests (for continuous group metrics); the paired comparisons are compiled in Table~\ref{tab:app_sig} (Appendix~\ref{app:sig}). For published leaderboard comparisons, we cite peer-reviewed or official leaderboard figures for 4B--70B models including GPT-4o, Claude-Sonnet-4, ArmoRM-8B~\cite{wang2024armorm}, Llama-3.1-8B-Instruct, Qwen3.5-9B/27B, GPT-OSS-20B, and Nemotron-70B.

Quantitative figures (Figures~\ref{fig:ablations}--\ref{fig:mrag}) are rendered in Matplotlib~3.11.2 directly from archived evaluation and paired-resampling reports, while the case-study and mechanism schematics (Figures~\ref{fig:case}, \ref{fig:framework}, and Figure~\ref{fig:comparison}) are generated as programmatic 2D vector SVGs rendered via Cairo and librsvg. The case study uses the saved forward-pass trace, and the RGB pixel hashes of its six photographs were checked against the model inputs.

\section{Results and Analysis}
\label{sec:results}

\subsection{Main Multi-Primitive Evolution and Benchmark Comparison}
\label{sec:main_results}

Table~\ref{tab:main_evolution} tracks five core metrics available across all seven 4B variants, while Table~\ref{tab:sota_external} positions \textbf{BiArbiter-4B} against published and official leaderboard references (full RewardBench~2, RuVerBench sub-dimension, MMBench TEST, paired significance, and variant diagnostic breakdowns appear in Appendices~\ref{app:rewardbench}--\ref{app:sig} and~\ref{app:evolution_diagnostics}).

\begin{table}[tbp]
\centering\tableformat
\caption{Seven 4B variants across four decision tasks. Bold scores mark column maxima. The last row gives the change from the frozen bidirectional base (A1).}
\label{tab:main_evolution}
\renewcommand{\arraystretch}{1.15}
\setlength{\tabcolsep}{2pt}
\begin{tabular*}{\linewidth}{@{\extracolsep{\fill}}lrrrrr@{}}
\toprule
 & \multicolumn{1}{c}{Routing} & \multicolumn{1}{c}{Ranking} & \multicolumn{2}{c}{Verification} & \multicolumn{1}{c@{}}{Choice} \\
\cmidrule(lr){2-2}\cmidrule(lr){3-3}\cmidrule(lr){4-5}\cmidrule(l){6-6}
Model variant & Macro-F1 & RB v1 & DR & AC & MM DEV \\
\midrule
A0: Causal base & 0.0864 & 84.65 & 85.15 & 58.35 & 88.31 \\
A1: Bidirectional base & 0.1739 & 86.28 & 85.59 & 57.98 & 90.00 \\
\addlinespace
A2: Causal LoRA & 0.3302 & 86.63 & 84.48 & 58.83 & 88.64 \\
A3: Bidirectional LoRA & 0.4327 & 87.67 & \textbf{86.05} & 57.13 & 90.00 \\
\addlinespace
Step 6: Balanced roster & 0.8394 & 87.93 & 85.26 & 61.74 & \textbf{90.85} \\
Step 7: Hard negatives & 0.8517 & 88.50 & 85.59 & 62.72 & 90.51 \\
\midrule
\rowcolor{biarbiterRow}
\textbf{BiArbiter-4B} & \textbf{0.8584} & \textbf{89.09} & 85.84 & \textbf{65.77} & 90.51 \\
$\Delta$ vs.\ A1 & $+0.6845$ & $+2.81$ & $+0.25$ & $+7.80$ & $+0.51$ \\
\bottomrule
\end{tabular*}
\par\smallskip
\begin{minipage}{\linewidth}\footnotesize
Routing uses InFoBench task macro-F1 (0--1). The other scores use a 0--100 scale: RewardBench v1 overall score, RuVerBench balanced accuracy on DeepResearch (DR) and AgenticCoding (AC), and MMBench DEV CircularEval on 590 training-disjoint groups. Step~6 and Step~7 correspond to Stage~2 and the Stage~3 midpoint ($\theta_{\mathrm{ref}}$), respectively. Differences ($\Delta$) are computed from full-precision values before rounding. Table~\ref{tab:evolution_diagnostics} reports empty dispatches and verification AUROCs.
\end{minipage}
\end{table}

\begin{table}[tbp]
\centering\tableformat
\caption{BiArbiter-4B and selected external references on preference ranking, rubric verification, and visual choice. All scores use a 0--100 scale.}
\label{tab:sota_external}
\renewcommand{\arraystretch}{1.12}
\setlength{\tabcolsep}{2pt}
\begin{tabular*}{\linewidth}{@{\extracolsep{\fill}}lrlr@{}}
\toprule
Benchmark or split & BiArbiter-4B & Reference model & Score \\
\midrule
\multicolumn{4}{@{}l}{\textit{Preference ranking}~\cite{lambert2025rewardbench,malik2026rewardbench2,wang2024armorm}} \\
RewardBench v1 & 89.09 & Nemotron-70B RM & 94.1 \\
 & & ArmoRM-Llama3-8B & 90.8 \\
 & & GPT-4o & 86.7 \\
\addlinespace
RewardBench 2 & 71.85 & Claude-Opus-4 & 76.5 \\
 & & INF-ORM-Llama3.1-70B & 76.5 \\
 & & Llama-3.1-8B-RM-RB2 & 71.9--72.8 \\
\midrule
\multicolumn{4}{@{}l}{\textit{RuVerBench verification (balanced accuracy)}~\cite{peng2026ruverbench,peng2026ruverbench_paper}} \\
DeepResearch & 85.84 & Gemini-3.1 Pro & 94.73 \\
 & & Qwen3.5-9B & 88.10 \\
\addlinespace
AgenticCoding & 65.77 & GPT-5.4 & 89.39 \\
 & & Qwen3.5-9B & 71.60 \\
\midrule
\multicolumn{4}{@{}l}{\textit{MMBench EN V1.1 visual choice (CircularEval)}} \\
DEV, full & 90.56 & Qwen3.5-4B (LiquidAI) & 86.50 \\
DEV, train-disjoint & 90.51 & & \\
TEST, train-disjoint & 88.45 & & \\
\bottomrule
\end{tabular*}
\par\smallskip
\begin{minipage}{\linewidth}\footnotesize
The full DEV split includes 702 of 1,292 groups overlapping training data. DEV, full reports the independent rerun (primary 90.33); RuVerBench rows report the primary run (reruns 85.88 and 65.18). Train-disjoint rows cover 590 DEV and 1,896 TEST groups without matched external references here. Tables~\ref{tab:app_rb2}, \ref{tab:app_ruver}, and~\ref{tab:app_mmbench} give full splits and reference rows.
\end{minipage}
\end{table}

Four headline findings emerge from Tables~\ref{tab:main_evolution} and~\ref{tab:sota_external}:
\begin{itemize}
    \item \textbf{Preference Ranking (RewardBench v1 \& 2):} \textbf{BiArbiter-4B} advances from 86.28\% (\texttt{A1}) to \textbf{89.09\%} on RewardBench v1 (+2.81\,pp, 95\% CI [+1.68, +3.92], $p=0.0005$; raw pair accuracy 89.95\%, McNemar $+141/-58$, $p<10^{-8}$), surpassing GPT-4o (86.7\%) by +2.39\,pp and approaching the dedicated 8B text reward model ArmoRM-8B (90.8\%). Stage-4 repair lifts \texttt{xstest-should-respond} from 80.00\% to 86.80\% (+6.80\,pp) and Chat Hard from 83.99\% to 85.09\% while holding Reasoning above 90\% (\texttt{hep-python} 98.17\%; Table~\ref{tab:app_rb1}). On RewardBench 2, it reaches \textbf{71.85}, exceeding published references GPT-4o (64.9), ArmoRM-8B (66.5), Claude-Sonnet-4 (71.2), and Skywork-Reward-Llama-3.1-8B-v0.2 (71.7); it gains +5.19\,pp over \texttt{A0} ($p=10^{-4}$) and is statistically tied with \texttt{A1} (72.55, $-0.70$\,pp, $p=0.487$), trading lower Ties ($-8.37$\,pp) and Factuality ($-5.68$\,pp) for higher Safety (+5.56\,pp), Focus (+2.83\,pp), and Math (+2.73\,pp) (Table~\ref{tab:app_rb2}).
    \item \textbf{Long-Horizon Rubric Verification (RuVerBench):} On DeepResearch, \textbf{BiArbiter-4B} attains \textbf{85.84\%} BAcc (rerun 85.88\%, AUROC \textbf{0.9442}), exceeding GPT-OSS-20B (85.6\%) and Llama-3.1-8B-Instruct (51.6\%). On AgenticCoding, it raises BAcc from 57.98\% to \textbf{65.77\%} (+7.80\,pp, 95\% CI [+1.39, +13.51], case-bootstrap $p=0.018$; rerun 65.18\%), surpassing Llama-3.1-8B-Instruct (61.7\%) as planning accuracy jumps from 63.46\% (\texttt{step7\_mid}) to 73.08\% and four-category macro negative recall climbs from 36.04\% to 55.37\% (Table~\ref{tab:app_ruver}).
    \item \textbf{Multimodal Visual Choice (MMBench DEV \& TEST):} On strictly unseen DEV (590 groups), \textbf{BiArbiter-4B} preserves \textbf{90.51\%} (zero forgetting vs.\ \texttt{step7\_mid}, +0.51\,pp over \texttt{A1}; 89.26\% on the 326 groups outside all v3 splits). On official MMBench TEST EN V1.1, it scores \textbf{88.45\%} across 1,896 strictly unseen groups (88.42\% across all 1,925 groups; Table~\ref{tab:app_mmbench}).
    \item \textbf{Multi-Label Routing (InFoBench Holdout):} Task Macro-F1 climbs from 0.0269 (frozen base, original prompt) / 0.1739 (roster) to \textbf{0.8584} ($p=0.0001$), eliminating empty dispatch (0/80) and resolving 7/8 linguistic/style-versus-content conflicts.
\end{itemize}

\subsection{Controlled Mechanism Ablations (Packages A1, A2, and A3)}
\label{sec:ablations}

\begin{figure}[tbp]
    \centering
    \includegraphics[width=0.98\linewidth]{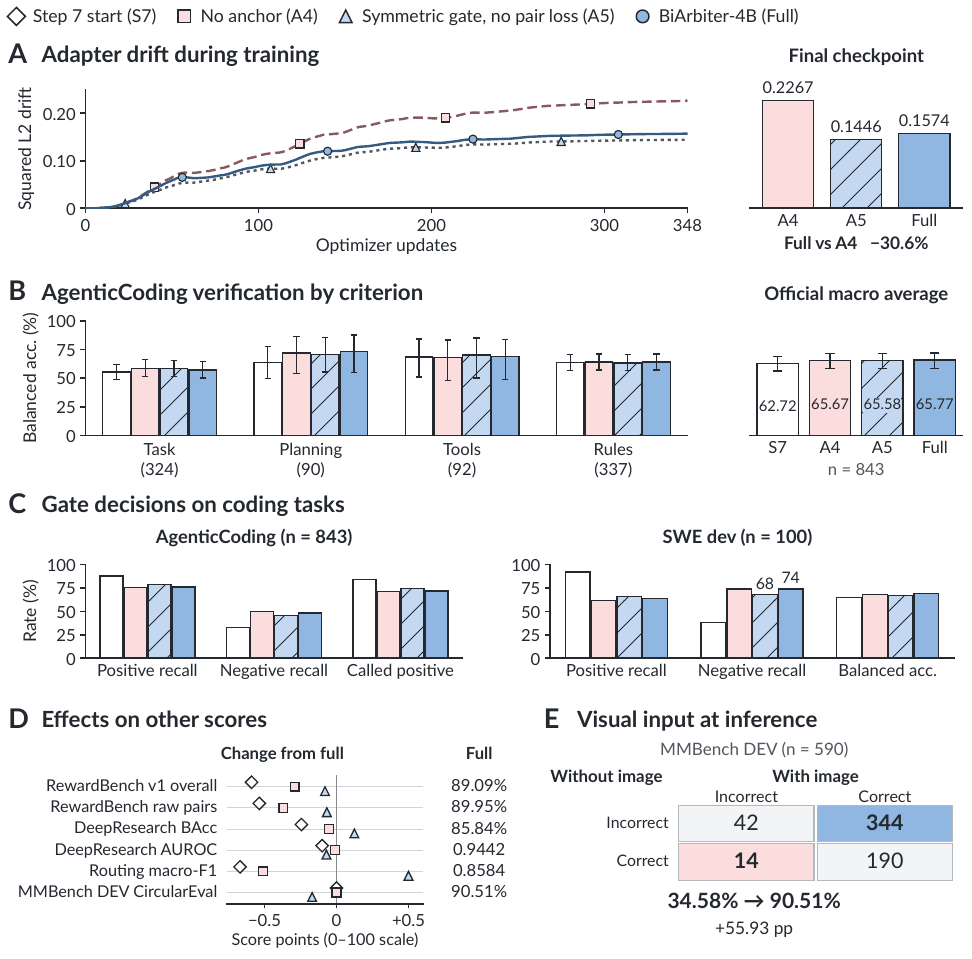}
    \caption{Matched Stage-4 ablations and visual input. Three variants share the Stage~3 midpoint (Step~7) initialization, seed, and 2,781 training sets. (A) Squared adapter drift across 348 updates, with final saved-weight distances at right. (B) AgenticCoding balanced accuracy with 95\% intervals from 10,000 paired resamples of 210 cases. (C) Gate decisions by class on AgenticCoding and SWE dev. (D) Differences from the full model on a 0--100 scale, with native scores at right. (E) Paired CircularEval outcomes with and without images on 590 training-disjoint MMBench DEV groups. Table~\ref{tab:supp_ablation} lists exact values.}
    \label{fig:ablations}
\end{figure}

Figure~\ref{fig:ablations} dissects the causal contribution of bidirectional attention, Stage-4 regularization, and visual grounding under matched budgets (full numeric records appear in Table~\ref{tab:supp_ablation}).

\paragraph{Causal vs.\ Bidirectional Attention (A1)}
Comparing \texttt{A0} with \texttt{A1} in Table~\ref{tab:main_evolution} (and Figure~\ref{fig:comparison}(C)) isolates the zero-parameter effect of unmasking the eight full-attention layers: RewardBench v1 rises +1.63\,pp (raw McNemar $p=0.048$), RewardBench 2 jumps +5.89\,pp (66.66 to 72.55), unseen MMBench rises +1.69\,pp (88.31\% to 90.00\%), DeepResearch AUROC rises from 0.9315 to 0.9392, and roster routing F1 doubles (0.0864 to 0.1739). Fine-tuning does not erase this advantage: after identical 255-step Round-1 LoRA training (\texttt{A2} vs.\ \texttt{A3}), the bidirectional model widens its lead on DeepResearch (+1.57\,pp, whereas causal fine-tuning regresses), RewardBench v1 (+1.04\,pp), unseen MMBench (+1.36\,pp), and routing F1 (+0.1025). Yet neither Round-1 arm solves AgenticCoding trajectory bias (58.83\% / 57.13\%), motivating Stage-4 calibration.

\paragraph{Single-Factor Training Ablations (A2)}
As shown in Figure~\ref{fig:ablations}(A)--(D), removing the $L_2$ parameter anchor (\texttt{A4}, $\lambda_{\mathrm{anc}}=0$) expands final squared LoRA drift $\|\theta - \theta_{\mathrm{ref}}\|_2^2$ by \textbf{+44.0\%} (0.1574 to 0.2267; anchoring cuts drift by 30.6\%) and degrades every downstream endpoint uniformly, including RewardBench v1 raw pair accuracy ($-0.37$\,pp, McNemar $p=0.0034$), routing F1 (0.8584 to 0.8533), AgenticCoding planning ($-1.3$\,pp), and SWE dev BAcc (69.0\% to 68.0\%). Removing asymmetric negative weighting and pairwise gate loss (\texttt{A5}) cuts SWE dev negative recall by $-6.0$\,pp (74.0\% to 68.0\%, BAcc 67.0\%), allows AgenticCoding positive prediction rate to rebound from 71.8\% to 74.6\% (planning $-2.3$\,pp), and dips unseen MMBench to 90.34\%. Because \texttt{A5} still reaches 65.58\% on AgenticCoding (and edges slightly higher on DR 85.96\% and routing F1 0.8634), these controls transparently separate the primary contribution of zero-overlap balanced SWE data (+2.86\,pp) from targeted negative-recall refinement (asymmetric pairwise gating) and cross-task drift control ($L_2$ anchoring).

\paragraph{Blind-Image Visual Grounding (A3)}
Figure~\ref{fig:ablations}(E) confirms genuine multimodal grounding: withholding images (\texttt{blind=True}) collapses strictly unseen MMBench DEV accuracy from \textbf{90.51\%} (534/590) to \textbf{34.58\%} (204/590; $-55.93$\,pp CircularEval, $+344/-14$ paired group verdicts, McNemar $p<10^{-10}$) and drops \texttt{v3\_holdout} (120 images) from 90.83\% to 37.50\%, ruling out text-only shortcuts.

\subsection{Measured Inference Cost (Package B1)}
\label{sec:efficiency}

\begin{figure}[tbp]
    \centering
    \includegraphics[width=0.98\linewidth]{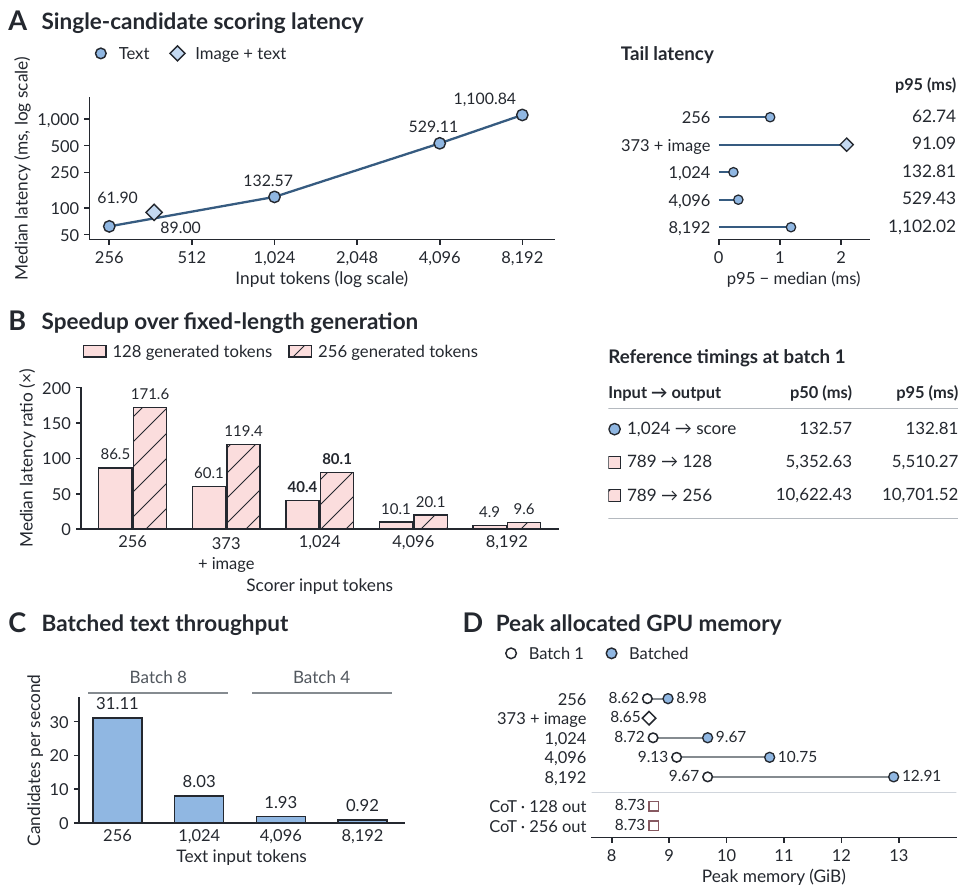}
    \caption{Inference cost on one A100 MIG \texttt{3g.40gb} slice (BF16, SDPA). (A) Median and 95th-percentile latency across repeated timings of fixed scoring inputs; the diamond denotes a synthetic 512-pixel image--text input. (B) Median causal generation time (789 input tokens) divided by median one-pass scoring time at batch size 1. (C) Candidate throughput at the indicated batch sizes; each text batch repeats equal-length inputs. (D) Peak allocated GPU memory for single candidates and measured text batches. Table~\ref{tab:supp_efficiency} lists all exact timings.}
    \label{fig:efficiency}
\end{figure}

Figure~\ref{fig:efficiency} summarizes single-candidate latency, batched text throughput, peak GPU memory, and speedup over causal chain-of-thought (CoT) judging on the same 4B backbone (full timing tables appear in Table~\ref{tab:supp_efficiency}).

At 256 tokens, \textbf{BiArbiter-4B} scores a candidate in 61.9\,ms (31.1 candidates/s at batch size 8, 8.62\,GiB single / 8.98\,GiB batched VRAM). At the 1,024-token reference length, one-pass scoring takes \textbf{132.6\,ms} versus 5,352.6\,ms (128-token CoT) and 10,622.4\,ms (256-token CoT) for the same KV-cached 4B backbone---a \textbf{40.4$\times$} to \textbf{80.1$\times$} speedup. Even 8,192-token multi-turn coding trajectories score in 1.10\,s under 9.67\,GiB single-candidate VRAM (12.91\,GiB at batch size 4), and 512-pixel vision-language inputs score in 89.0\,ms (8.65\,GiB).

\subsection{Probability Calibration and Selective Prediction (Package B2)}
\label{sec:calibration}

\begin{figure}[tbp]
    \centering
    \includegraphics[width=0.98\linewidth]{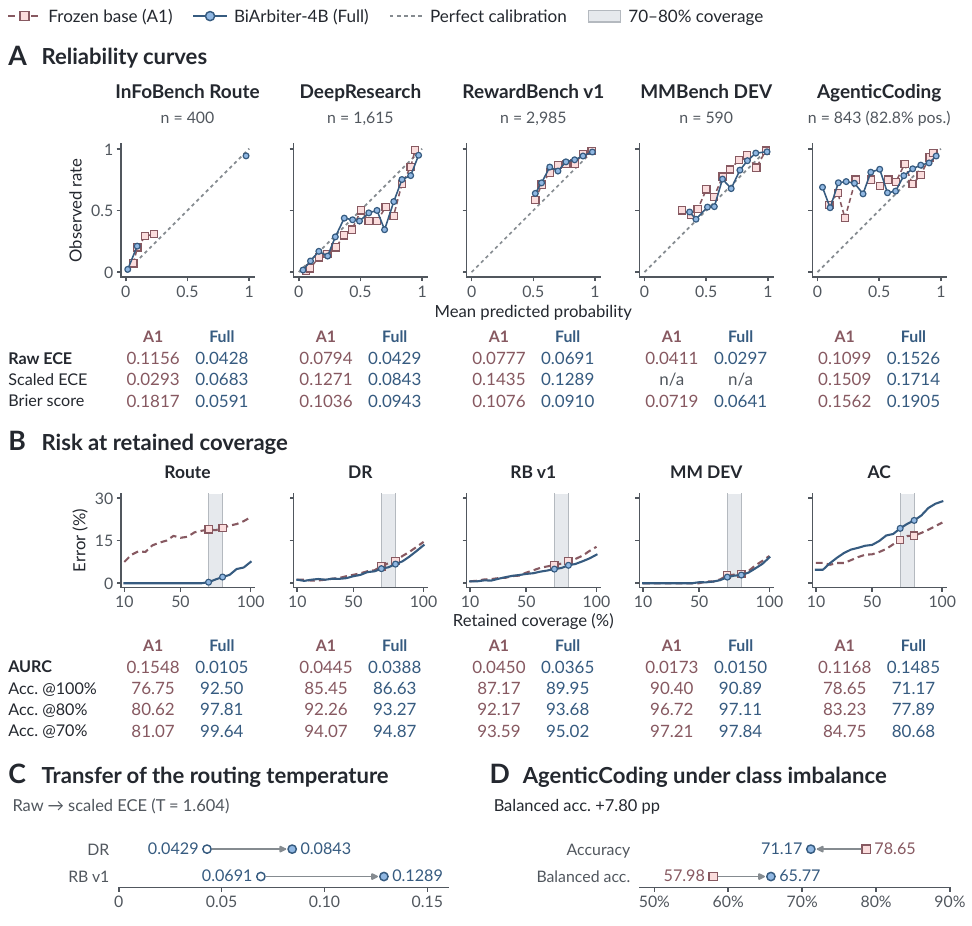}
    \caption{Calibration and selective prediction for the frozen base (A1) and BiArbiter-4B. (A) Reliability curves over 15 equal-width bins (points require $\ge 10$ observations; ECE includes all non-empty bins). (B) Selective risk--coverage curves with 70--80\% coverage shaded; AURC denotes area under the risk--coverage curve. (C) Applying the routing dev-fitted temperature ($T_{\mathrm{dev}}=1.604$) worsens ECE on DeepResearch and RewardBench v1. (D) On class-imbalanced AgenticCoding (82.8\% positive), overall accuracy drops while macro balanced accuracy rises by $+7.80$\,pp. Table~\ref{tab:supp_calibration} lists exact values.}
    \label{fig:calibration}
\end{figure}

Figure~\ref{fig:calibration} evaluates native 15-bin Expected Calibration Error (ECE), cross-task temperature scaling, Brier score, and selective risk--coverage (AURC) across all five benchmark splits (complete tabular figures appear in Table~\ref{tab:supp_calibration}).

Relative to the frozen bidirectional base, \textbf{BiArbiter-4B} improves raw ECE (0.0297--0.0691), Brier score, and AURC across all four balanced or standard splits (InFoBench routing, RuVerBench-DR, RewardBench v1, and unseen MMBench DEV), and achieves the lowest AURC across all four evaluated checkpoints (frozen base, Stage~2, Stage~3 midpoint, and BiArbiter-4B: 0.0105, 0.0388, 0.0365, and 0.0150; Table~\ref{tab:supp_calibration}). On InFoBench routing, AURC drops 14.7$\times$ (0.1548 to \textbf{0.0105}), reaching 97.81\% accuracy at 80\% coverage and 99.64\% at 70\% coverage; on DeepResearch and RewardBench v1, escalating the lowest-confidence 20--30\% of judgments lifts 4B accuracy to 93.27--95.02\%. Crucially, Figure~\ref{fig:calibration}(C) shows that transferring a routing-dev temperature ($T_{\mathrm{dev}}=1.604$) degrades ECE on already-calibrated DeepResearch (0.0429 to 0.0843) and RewardBench v1 (0.0691 to 0.1289), confirming that anchored training yields natively usable confidence scores without post-hoc rescaling. Conversely, Figure~\ref{fig:calibration}(D) transparently documents the expected class-imbalance trade-off on RuVerBench-AC (82.8\% positive, 698/843 checks): asymmetric negative weighting delivers the +7.80\,pp macro BAcc gain while trading off majority-weighted raw accuracy (78.65\% to 71.17\%), raw ECE (0.1099 to 0.1526), and AURC (0.1168 to 0.1485).

\subsection{Downstream Closed-Loop Utility (Packages C1 and C2)}
\label{sec:closed_loop}

\begin{figure}[tbp]
    \centering
    \includegraphics[width=0.98\linewidth]{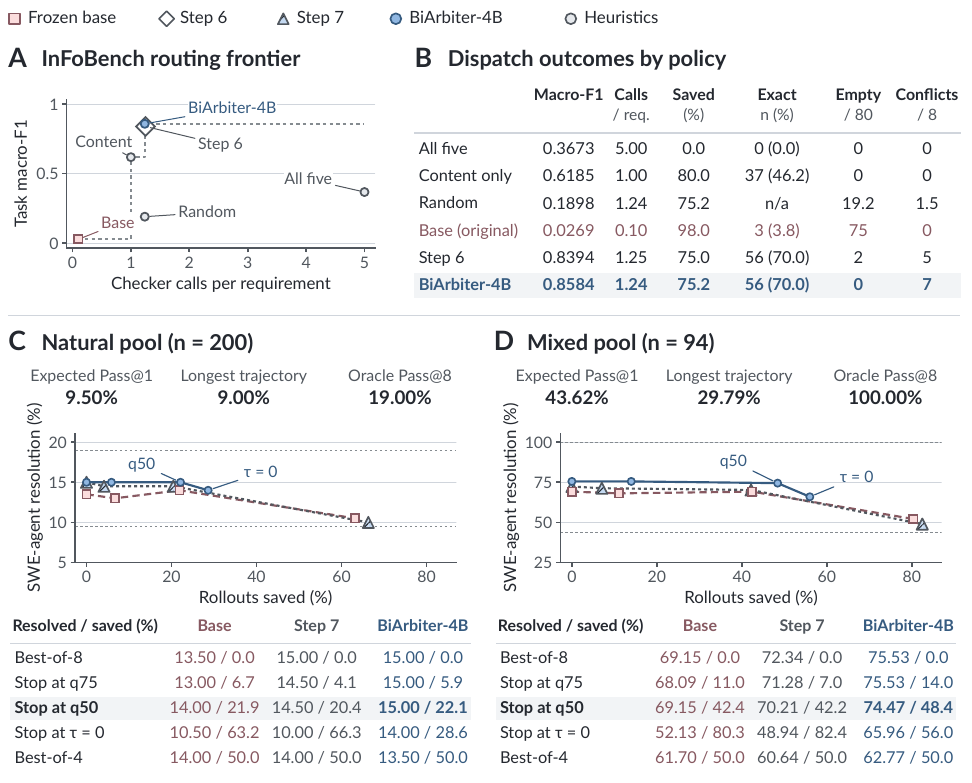}
    \caption{Checker routing and selection of held-out SWE-agent trajectories with recorded unit-test outcomes (shards 04--05, $K=8$, zero shared issue IDs with train/dev; 34 IDs shared between the separately reported pools). (A, B) Routing frontier and dispatch reliability across six policies on 22 InFoBench tasks (80 requirements). (C, D) Natural and mixed SWE-agent pools connecting Best-of-8 with sequential early stopping at development-fitted $q_{75}$, $q_{50}$, and $\tau=0$; at $q_{50}$, BiArbiter-4B uses 1,246 of 1,600 rollouts in the natural pool and 388 of 752 in the mixed pool, and dotted lines mark expected Pass@1 and oracle Pass@8. Table~\ref{tab:supp_downstream} lists exact values.}
    \label{fig:downstream}
\end{figure}

Figure~\ref{fig:downstream} evaluates \textbf{BiArbiter-4B} inside live execution loops for cost-aware checker routing and held-out software-engineering trajectory selection (full numeric tables appear in Table~\ref{tab:supp_downstream}).

\paragraph{Cost-Aware Checker Routing (C1)}
Invoking all five checkers wastes 76.3\% of calls on irrelevant checks (Micro Precision 0.237, Macro-F1 0.3673), whereas Content-Only misses 46.3\% of required checks (Macro-F1 0.6185, 0/8 conflicts). As shown in Figure~\ref{fig:downstream}(A)--(B), \textbf{BiArbiter-4B} sits on the Pareto frontier: it averages just \textbf{1.238} checkers per requirement (\textbf{75.2\% compute savings} vs.\ Always-All), lifting task Macro-F1 to \textbf{0.8584} ($p=0.0001$ vs.\ original-prompt \texttt{base}; 0.828 micro precision / 0.863 micro recall), 70.0\% exact multi-label match, 0/80 empty dispatches, and 7/8 conflict resolution (exact McNemar $p=0.0156$ vs.\ \texttt{base}).

\paragraph{Real Unit-Test Reranking and Early Stopping on Unseen GitHub Issues (C2)}
On strictly disjoint \texttt{SWE-agent} shards 04--05 (0 shared \texttt{instance\_id}s with training or development; Natural $n=200$ and Mixed $n=94$ reported separately), \textbf{BiArbiter-4B} Best-of-8 lifts mixed-pool resolution from 43.62\% (expected Pass@1) and 29.79\% (longest trajectory) to \textbf{75.53\%} (+31.91\,pp over Pass@1, $p=0.0001$; +6.38\,pp over \texttt{base} 69.15\%) and natural-pool resolution from 9.50\% to \textbf{15.00\%} (+5.50\,pp, $p=0.0001$, candidate AUROC \textbf{0.8721}). Figure~\ref{fig:downstream}(C)--(D) further demonstrates why calibration governs sequential early stopping: at the uncalibrated zero threshold ($\tau=0$), frozen \texttt{base} stops prematurely after 1.57 rollouts/issue and collapses to 52.13\%, whereas calibrated \textbf{BiArbiter-4B} sustains \textbf{65.96\%} (+13.83\,pp, McNemar $p=0.0146$) at 56.0\% rollout savings. Under the dev-frozen positive median ($\tau_{q50}=0.259$), \textbf{BiArbiter-4B} reaches \textbf{74.47\%} on mixed issues (within 1.06\,pp of full Best-of-8 while saving \textbf{48.4\%} of rollouts) and \textbf{15.00\%} on natural issues (matching full Best-of-8 while saving \textbf{22.1\%} of rollouts).

\subsection{Zero-Train Cross-Application Domain Generalization (Package C3 \& External Transfer)}
\label{sec:cross_domain}

\begin{figure}[tbp]
    \centering
    \includegraphics[width=0.98\linewidth]{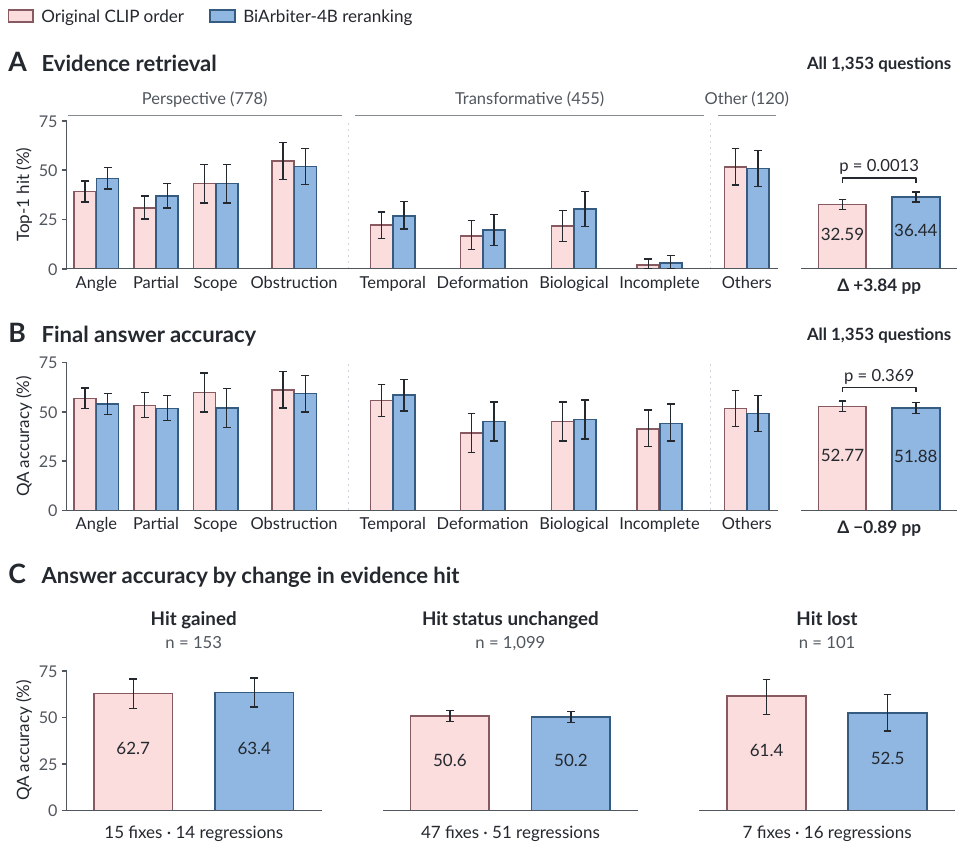}
    \caption{Evidence reranking with a fixed answer scorer on 1,353 MRAG-Bench questions. (A, B) Top-1 evidence hit rate and final answer accuracy by scenario and overall. (C) Answer accuracy grouped by change in evidence hit status; Figure~\ref{fig:case} illustrates one of 15 answer corrections among the 153 questions that gained a hit. Whiskers show 95\% intervals from 10,000 paired question-level bootstrap resamples; brackets report two-sided exact McNemar tests.}
    \label{fig:mrag}
\end{figure}

Table~\ref{tab:transfer} evaluates the identical frozen \textbf{BiArbiter-4B} checkpoint across three unseen visual, document, and industrial domains plus two held-out constraint and surface-bias stress tests, while Figure~\ref{fig:mrag} unpacks all nine MRAG-Bench scenarios and connects directly to the motivating case in Figure~\ref{fig:case}.

\begin{table}[tbp]
\centering\tableformat
\setlength{\tabcolsep}{2.2pt}
\caption{Zero-train transfer to held-out applications and stress tests with identical frozen weights (no domain adapter or test-domain threshold tuning).}
\label{tab:transfer}
\begin{tabular*}{\linewidth}{@{\extracolsep{\fill}}lrlrrrrr@{}}
\toprule
Benchmark & $n$ & Primary metric & Simple & Base & BiArbiter & $\Delta_{\mathrm{Simple}}$ & $\Delta_{\mathrm{Base}}$ \\
\midrule
MRAG all & 1,353 & Top-1 hit (\%) & 32.59 & 38.36 & 36.44 & $+3.84^{\dagger}$ & $-1.92$ \\
MRAG mixed & 731 & Top-1 hit (\%) & 56.91 & 67.58 & 64.02 & $+7.11^{\dagger}$ & $-3.56$ \\
MRAG transformative & 455 & QA acc.\ (\%) & 46.37 & 50.55 & 49.45 & $+3.08^{\ddagger}$ & $-1.10$ \\
MP-DocVQA & 3,554 & Page top-1 (\%) & 28.67 & 75.30 & 74.14 & $+45.47$ & $-1.15$ \\
MVTec-AD & 198 & Macro AUROC & 0.500 & 0.871 & 0.868 & $+0.368$ & $-0.003$ \\
IFBench strict & 295 & AUROC & 0.500 & 0.705 & 0.730 & $+0.230$ & $+0.025$ \\
IFBench loose & 295 & AUROC & 0.500 & 0.723 & 0.742 & $+0.242$ & $+0.018$ \\
RubricBench & 1,146 & Pair acc.\ (\%) & 44.85 & 52.09 & 49.04 & $+4.19^{\ast}$ & $-3.05$ \\
\bottomrule
\end{tabular*}
\par\smallskip
\begin{minipage}{\linewidth}\footnotesize
Simple denotes original CLIP order (MRAG), first page (MP-DocVQA), chance AUROC (MVTec-AD, IFBench), or the length heuristic (RubricBench). Differences ($\Delta$) are computed from full-precision values before rounding, reported in percentage points for percentage metrics and raw AUROC otherwise. Exact two-sided McNemar tests vs.\ Simple: $^{\dagger}p=0.0013$, $^{\ddagger}p=0.0704$, $^{\ast}p=0.0485$. Secondary metrics (MRR, NDCG@5, BAcc), subgroup breakdowns, and SHA-256 overlap audits appear in Table~\ref{tab:app_cross}.
\end{minipage}
\end{table}

Two complementary lessons stand out from Table~\ref{tab:transfer} and Figure~\ref{fig:mrag}:
\begin{enumerate}
    \item \textbf{Strong Zero-Shot Utility Across Unseen Domains:} Without RAG, document, or industrial training data, frozen \textbf{BiArbiter-4B} lifts MRAG-Bench visual evidence Top-1 hit by \textbf{+3.84\,pp} overall (32.59\% to 36.44\%, $p=0.0013$) and \textbf{+7.11\,pp} on mixed retrieval (56.91\% to 64.02\%), raising \texttt{Transformative} QA by +3.08\,pp (46.37\% to 49.45\%). On MP-DocVQA, it routes multipage evidence at \textbf{74.14\%} Top-1 (+45.47\,pp over first-page; MRR 0.839). On MVTec-AD, it reaches \textbf{0.868} macro AUROC (\texttt{bottle} 0.884, \texttt{metal\_nut} 0.851), and on unseen IFBench constraint gating it beats \texttt{base} on both strict (+0.025 AUROC) and loose (+0.018) splits (Table~\ref{tab:app_cross}).
    \item \textbf{Quantified Retention--Adaptation Trade-Off and Retrieval-to-QA Anatomy:} Relative to unfine-tuned \texttt{base}, anchored adaptation incurs a modest, systematic prior drift across unseen visual domains: $-0.003$ AUROC on MVTec-AD, $-1.15$\,pp on MP-DocVQA, and $-1.92$\,pp on MRAG-Bench Top-1 (while yielding sharper selective QA confidence on mixed MRAG-Bench at 80--90\% coverage: 61.37\%/60.33\% vs.\ 60.51\%/58.97\%). Furthermore, Figure~\ref{fig:mrag} explains why higher overall retrieval hit rate did not raise aggregate MRAG-Bench QA (52.77\% to 51.88\%, $p=0.369$): while \texttt{Transformative} shifts gain both hits (+4.40\,pp) and QA (+3.08\,pp), \texttt{Perspective} shifts gain hits (+4.24\,pp) yet dip in QA ($-2.96$\,pp) when dual-image distractors interfere with questions already answerable from the query image alone. These figures quantify the exact 0.3--1.9\,pp prior-drift cost required to convert a base model that fails at multi-label routing (F1 0.027) and agent early stopping (52.1\%) into a calibrated unified arbiter.
\end{enumerate}

\section{Discussion, Error Anatomy, and Limitations}
\label{sec:discussion}

Synthesizing all 11 benchmarks highlights five concrete capability boundaries that inform safe deployment and future design:

\paragraph{1. Cross-Framework Agent Protocol Shift}
Although \textbf{BiArbiter-4B} reaches 75.53\% Best-of-8 resolution on unseen \texttt{SWE-agent} issues sharing the CLI scaffold, its ROC-AUC on \texttt{RuVerBench-AC} (\texttt{Claude-Code} multi-agent traces with specialized MCP and sub-agent schemas) sits at 0.7106 (below \texttt{base} 0.7437 despite +7.80\,pp higher BAcc), trailing Qwen3.5-9B (71.6\%) and GPT-5.4 (89.4\%). Calibrating verifiers across heterogeneous agent scaffolds without overfitting one tool syntax remains an open challenge.

\paragraph{2. Pairwise Margin Sharpening vs.\ Ties}
On RewardBench 2, pairwise margin training boosts Safety (+5.56\,pp), Focus (+2.83\,pp), and Math (+2.73\,pp) over \texttt{base}, but lowers Ties by $-8.37$\,pp because margin pushing penalizes small score gaps between equally valid responses. Explicit tie-preservation regularization---penalizing $|z_i - z_j|$ on equivalent pairs---offers a direct remedy.

\paragraph{3. Parametric Ceiling on Implicit Constraints and Surface Bias}
On RewardBench 2 Precise IF (44.38\%) and RubricBench without explicit rubrics (49.04\% vs.\ 44.85\% length heuristic, $p=0.0485$), a single 4B forward pass remains bounded on dense implicit constraints and exact counting rules, corroborating the \emph{Rubric Gap}~\cite{zhou2026rubricbench}. Rather than undermining the framework, this boundary reinforces our system architecture: verifiable constraints should be routed to deterministic checkers (Figure~\ref{fig:downstream}(A)--(B)), while low-confidence complex comparisons should escalate via selective abstention (Figure~\ref{fig:calibration}(B)).

\paragraph{4. Fine-Grained Visual Reasoning Sub-Skills}
On strictly unseen MMBench TEST (1,896 groups, 88.45\% overall), errors cluster in \texttt{physical\_property\_reasoning} (73.97\%, 54/73), \texttt{physical\_relation} (76.00\%, 57/75), and \texttt{image\_quality} (76.92\%, 90/117), whereas \texttt{ocr} (97.78\%), \texttt{image\_topic} (97.78\%), and \texttt{attribute\_comparison} (95.95\%) approach ceiling (Table~\ref{tab:app_mmbench}).

\paragraph{5. Scope of Reproducibility}
Our results report one locked training seed with group-level paired bootstrap intervals across 11 benchmarks, an independent full rerun confirming near-deterministic BF16 inference, and a merged BF16 release check on RewardBench 2 (71.52 vs.\ 71.85 for dynamic LoRA; Table~\ref{tab:app_rb2}). Quantifying multi-seed variance and extending anchored bidirectional scoring to 8B--9B hybrid backbones are direct avenues for future work.

\section{Conclusion}
\label{sec:conclusion}

We introduced \textbf{BiArbiter-4B}, a unified 4B multimodal candidate scorer combining one-pass bidirectional hybrid-attention readout with anchored multi-stage calibration to serve routing, ranking, verification gating, and visual choice from one frozen LoRA checkpoint. Matched ablations establish distinct causal gains from bidirectional unmasking, $L_2$ parameter anchoring ($-30.6\%$ squared weight drift, RewardBench $p=0.0034$), asymmetric negative gate supervision (+6.0\,pp SWE negative recall), and real visual grounding (+55.93\,pp vs.\ blind-image). Running 40.4$\times$--80.1$\times$ faster than same-backbone CoT judging, \textbf{BiArbiter-4B} saves 75.2\% checker calls at 0.8584 Macro-F1, resolves 74.47\% of disjoint coding issues at 48.4\% fewer rollouts via early stopping, and transfers zero-shot across Visual-RAG, multipage document routing, and industrial anomaly inspection. Coupled with transparent reporting of its calibration trade-offs and capability boundaries, \textbf{BiArbiter-4B} offers a practical, reproducible foundation for low-latency knowledge-based decision systems; future work will address cross-scaffold agent calibration, tie-preserving margin objectives, and multi-seed scaling to larger hybrid backbones.


\section*{Reproducibility, Data Availability, and Declarations}
\paragraph{Data and Code Availability}
All external evaluation benchmarks (RewardBench v1/2, RuVerBench, MMBench EN V1.1, MRAG-Bench, MP-DocVQA, MVTec-AD, IFBench, RubricBench, InFoBench, HelpSteer2, and \texttt{nebius/\allowbreak SWE-agent-\allowbreak trajectories}) are publicly available from their respective official releases. Trained LoRA weights, zero-overlap split manifests, evaluation scripts, and raw per-item prediction logs will be released publicly upon formal journal publication.

\paragraph{Competing Interests}
The authors declare that they have no known competing financial interests or personal relationships that could have appeared to influence the work reported in this paper.

\paragraph{AI-Assisted Tools}
OpenAI Codex (GPT-6, OpenAI) assisted with code verification, language editing, figure layout and plotting scripts. An early exploratory layout of Figure~\ref{fig:framework} referenced GPT Image 2.6 Sunburst before the figure was redrawn as a programmatic 2D vector schematic. The plotted values come from the recorded evaluations, and the photographs in Figure~\ref{fig:case} come from MRAG-Bench.

\FloatBarrier
\setlength{\bibsep}{2.5pt plus 0.5pt}
\bibliographystyle{unsrtnat}
\bibliography{references}

\clearpage
\appendix
\renewcommand{\thetable}{\thesection.\arabic{table}}
\renewcommand{\thefigure}{\thesection.\arabic{figure}}
\makeatletter
\@addtoreset{table}{section}
\@addtoreset{figure}{section}
\makeatother
\renewcommand{\theHtable}{\thesection.\arabic{table}}
\renewcommand{\theHfigure}{\thesection.\arabic{figure}}

\section{Detailed RewardBench v1 and RewardBench 2 Breakdowns}
\label{app:rewardbench}

Tables~\ref{tab:app_rb1} and~\ref{tab:app_rb2} provide the full section-level and diagnostic-subset breakdowns for RewardBench v1 (2,985 pairs, 0 truncations) and RewardBench 2 (1,865 prompts / 8,977 candidates, 0 truncations), alongside published reference baselines~\cite{lambert2025rewardbench,malik2026rewardbench2}. On RewardBench v1, Stage-4 targeted repair specifically lifts \texttt{xstest-should-respond} (+6.80\,pp over \texttt{step7\_mid}) and \texttt{llmbar-adver-GPTOut} (+4.26\,pp) without losing safety refusal accuracy (\texttt{xstest-should-refuse} 96.10\%) or coding precision (\texttt{hep-python} 98.17\%). On RewardBench 2, \textbf{BiArbiter-4B} matches or surpasses 8B--70B and frontier references on Focus (95.56) and Precise IF (44.38 vs.\ 41.9 for Claude-Opus-4 and INF-ORM-70B), while revealing the margin-sharpening trade-off on Ties (78.81 vs.\ 87.18 for frozen \texttt{base}).

\begin{table}[tbp]
\centering\tableformat
\setlength{\tabcolsep}{2pt}
\caption{RewardBench v1 section scores and diagnostic accuracy across training stages (2,985 pairs, zero truncations).}
\label{tab:app_rb1}
\begin{tabular*}{\linewidth}{@{\extracolsep{\fill}}lrrrrr@{}}
\toprule
Score (0--100) & A0 & A1 & Step 6 & Step 7 & BiArbiter \\
\midrule
\multicolumn{6}{@{}l}{\textit{Aggregate and section scores}} \\
Weighted overall & 84.65 & 86.28 & 87.93 & 88.50 & \textbf{89.09} \\
Raw accuracy & 86.13 & 87.17 & 88.61 & 89.41 & \textbf{89.95} \\
Chat & 88.27 & 89.94 & 93.85 & \textbf{94.69} & 94.41 \\
Chat Hard & 83.77 & \textbf{85.96} & 85.31 & 83.99 & 85.09 \\
Safety & 78.92 & 82.03 & 83.78 & 85.14 & \textbf{86.76} \\
Reasoning & 87.66 & 87.19 & 88.77 & \textbf{90.18} & 90.12 \\
\addlinespace
\multicolumn{6}{@{}l}{\textit{Diagnostic subsets}} \\
XSTest should respond & \textbf{89.60} & 86.40 & 81.20 & 80.00 & 86.80 \\
XSTest should refuse & 92.86 & 95.45 & 95.45 & \textbf{96.10} & \textbf{96.10} \\
Do-Not-Answer & 64.71 & 69.85 & 72.06 & \textbf{73.53} & 70.59 \\
LLMBar GPTOut & 80.85 & 78.72 & \textbf{85.11} & 78.72 & 82.98 \\
HEP Python & 94.51 & 93.90 & 95.12 & 96.95 & \textbf{98.17} \\
\bottomrule
\end{tabular*}
\par\smallskip
\begin{minipage}{\linewidth}\footnotesize
A0 and A1 are the causal and bidirectional bases. Step 7 uses the intermediate checkpoint. BiArbiter denotes BiArbiter-4B. Bold marks the highest score in each row.
\end{minipage}
\end{table}

\begin{table}[tbp]
\centering\tableformat
\setlength{\tabcolsep}{2pt}
\caption{RewardBench 2 scores under the official rules for 1,865 prompts and 8,977 candidates (zero local truncations). Published scores follow the benchmark report~\cite{malik2026rewardbench2}.}
\label{tab:app_rb2}
\begin{tabular*}{\linewidth}{@{\extracolsep{\fill}}lrrrr@{}}
\toprule
Model & Overall & Focus & Safety & Ties \\
\midrule
\multicolumn{5}{@{}l}{\textit{Published references}} \\
Claude-Opus-4 & 76.5 & 86.2 & 89.5 & 83.7 \\
INF-ORM-70B & 76.5 & 90.3 & 96.4 & 86.2 \\
Llama-3.1-8B-RM-RB2 & 72.8 & 90.7 & 89.6 & 76.4 \\
Skywork-8B-v0.2 & 71.7 & -- & 94.2 & -- \\
ArmoRM-Llama3-8B & 66.5 & -- & -- & -- \\
GPT-4o & 64.9 & -- & -- & -- \\
\addlinespace
\multicolumn{5}{@{}l}{\textit{Local 4B evaluations}} \\
A0 causal base & 66.66 & 89.29 & 63.33 & 77.66 \\
A1 bidirectional base & \textbf{72.55} & 92.73 & 73.33 & \textbf{87.18} \\
BiArbiter-4B (LoRA) & 71.85 & \textbf{95.56} & \textbf{78.89} & 78.81 \\
BiArbiter-4B (merged) & 71.52 & 95.35 & 78.44 & 78.02 \\
\bottomrule
\end{tabular*}
\par\medskip
\begin{tabular*}{\linewidth}{@{\extracolsep{\fill}}lrrr@{}}
\toprule
Model & Math & Factuality & Precise IF \\
\midrule
\multicolumn{4}{@{}l}{\textit{Published references}} \\
Claude-Opus-4 & 74.9 & 82.7 & 41.9 \\
INF-ORM-70B & 69.9 & 74.1 & 41.9 \\
Llama-3.1-8B-RM-RB2 & 61.7 & 74.3 & 44.4 \\
Skywork-8B-v0.2 & 60.1 & 69.7 & 40.6 \\
ArmoRM-Llama3-8B & -- & -- & -- \\
GPT-4o & -- & -- & -- \\
\addlinespace
\multicolumn{4}{@{}l}{\textit{Local 4B evaluations}} \\
A0 causal base & 66.12 & 61.05 & 42.50 \\
A1 bidirectional base & 66.12 & \textbf{70.32} & \textbf{45.63} \\
BiArbiter-4B (LoRA) & \textbf{68.85} & 64.63 & 44.38 \\
BiArbiter-4B (merged) & 68.31 & 64.63 & 44.38 \\
\bottomrule
\end{tabular*}
\par\smallskip
\begin{minipage}{\linewidth}\footnotesize
Bold marks the highest score among local 4B evaluations in each column. Dashes indicate unreported reference scores. LoRA denotes the dynamically loaded adapter.
\end{minipage}
\end{table}

\FloatBarrier
\section{Detailed RuVerBench Sub-Dimension Leaderboard Comparison}
\label{app:ruverbench}

Table~\ref{tab:app_ruver} reports all eight semantic sub-dimensions of RuVerBench (DeepResearch: Format, Numbers, Logic, Facts; AgenticCoding: Task, Planning, Tools, Rules) scored with the official \texttt{compute\_main\_leaderboard.py} evaluator, alongside representative models from the official RuVerBench main leaderboard~\cite{peng2026ruverbench,peng2026ruverbench_paper}.

\begin{table}[tbp]
\centering\tableformat
\setlength{\tabcolsep}{2pt}
\caption{Balanced accuracy (0--100) by RuVerBench dimension for official leaderboard models and local 4B checkpoints.}
\label{tab:app_ruver}
\begin{tabular*}{\linewidth}{@{\extracolsep{\fill}}lrrrrr@{}}
\toprule
DeepResearch & Overall & Format & Numbers & Logic & Facts \\
\midrule
Gemini-3.1 Pro$^\ast$ & 94.7 & 95.6 & 93.8 & 95.7 & 93.8 \\
GPT-5.4 & 91.4 & 95.1 & 93.3 & 86.4 & 91.0 \\
Qwen3.5-27B & 89.8 & 92.9 & 92.2 & 86.0 & 88.3 \\
Qwen3.5-9B & 88.1 & 91.7 & 90.6 & 83.9 & 86.3 \\
GPT-OSS-20B & 85.6 & 91.9 & 85.9 & 82.1 & 82.3 \\
Llama-3.1-8B-Inst & 51.6 & 49.7 & 51.3 & 53.4 & 52.2 \\
\midrule
A0 causal & 85.15 & 90.93 & 85.20 & \textbf{82.21} & 82.26 \\
A1 bidirectional & 85.59 & 91.82 & \textbf{85.79} & 81.02 & 83.73 \\
Step 6 & 85.26 & 91.26 & 84.57 & 81.47 & 83.76 \\
Step 7 & 85.59 & 91.32 & 85.30 & 81.74 & 84.01 \\
BiArbiter-4B (primary) & 85.84 & \textbf{92.08} & 84.38 & 82.05 & 84.84 \\
BiArbiter-4B (rerun) & \textbf{85.88} & \textbf{92.08} & 84.38 & 82.05 & \textbf{85.04} \\
\bottomrule
\end{tabular*}
\par\medskip
\begin{tabular*}{\linewidth}{@{\extracolsep{\fill}}lrrrrr@{}}
\toprule
AgenticCoding & Overall & Task & Planning & Tools & Rules \\
\midrule
Gemini-3.1 Pro & 86.5 & 84.7 & 83.7 & 90.3 & 87.3 \\
GPT-5.4$^\ast$ & 89.4 & 86.4 & 92.6 & 89.7 & 88.9 \\
Qwen3.5-27B & 74.0 & 77.6 & 83.3 & 64.7 & 70.2 \\
Qwen3.5-9B & 71.6 & 71.5 & 74.0 & 74.0 & 66.8 \\
GPT-OSS-20B & 72.3 & 80.0 & 80.1 & 59.2 & 69.8 \\
Llama-3.1-8B-Inst & 61.7 & 56.8 & 62.8 & 72.0 & 55.2 \\
\midrule
A0 causal & 58.35 & \textbf{67.34} & 43.59 & 60.55 & 61.94 \\
A1 bidirectional & 57.98 & 62.73 & 46.15 & 56.85 & \textbf{66.17} \\
Step 6 & 61.74 & 60.43 & 57.37 & 64.70 & 64.45 \\
Step 7 & 62.72 & 55.23 & 63.46 & 68.41 & 63.79 \\
BiArbiter-4B (primary) & \textbf{65.77} & 57.11 & \textbf{73.08} & \textbf{68.86} & 64.05 \\
BiArbiter-4B (rerun) & 65.18 & 57.11 & 71.79 & \textbf{68.86} & 62.94 \\
\bottomrule
\end{tabular*}
\par\smallskip
\begin{minipage}{\linewidth}\footnotesize
DeepResearch covers 284 cases (1,615 rubrics) and AgenticCoding covers 210 cases (843 checks). The first six rows of each panel are official scores. Asterisks mark leaderboard leaders. Bold marks local maxima.
\end{minipage}
\end{table}

\FloatBarrier
\section{MMBench EN V1.1 Overlap Audit and 20-Category Breakdown}
\label{app:mmbench}

Table~\ref{tab:app_mmbench} documents the exact split audit on MMBench DEV EN V1.1 (confirming 0/590 group flips across two independent full reruns on the strictly unseen split) together with all 20 visual sub-skills of \textbf{BiArbiter-4B} on the strictly unseen MMBench TEST EN V1.1 partition (1,896 groups / 7,193 rows).

\begin{table}[tbp]
\centering\tableformat
\setlength{\tabcolsep}{2pt}
\caption{BiArbiter-4B on MMBench English V1.1. Category scores cover the 1,896 training-disjoint TEST groups.}
\label{tab:app_mmbench}
\begin{tabular*}{\linewidth}{@{\extracolsep{\fill}}lrrr@{}}
\toprule
DEV partition & Groups & Primary (\%) & Rerun (\%) \\
\midrule
All & 1,292 & 90.33 & 90.56 \\
Train overlap & 702 & 90.17 & 90.60 \\
Train-disjoint & 590 & 90.51 & 90.51 \\
Outside all v3 splits & 326 & 89.26 & 89.26 \\
\bottomrule
\end{tabular*}
\par\medskip
\begin{tabular*}{\linewidth}{@{\extracolsep{\fill}}lrrr@{}}
\toprule
TEST partition & Total & Correct & Accuracy (\%) \\
\midrule
All & 1,925 & 1,702 & 88.42 \\
Train overlap & 29 & 25 & 86.21 \\
Train-disjoint & 1,896 & 1,677 & 88.45 \\
Train-disjoint (vanilla) & 7,193 & 6,365 & 88.49 \\
\bottomrule
\end{tabular*}
\par\medskip
\begin{tabular*}{\linewidth}{@{\extracolsep{\fill}}lrrlrr@{}}
\toprule
Category & $n/N$ & \% & Category & $n/N$ & \% \\
\midrule
OCR & 88/90 & 97.78 & Image style & 81/92 & 88.04 \\
Image topic & 88/90 & 97.78 & Nature relation & 78/89 & 87.64 \\
Attribute comp. & 71/74 & 95.95 & Spatial rel. & 65/75 & 86.67 \\
Func. reasoning & 85/89 & 95.51 & Struct. image text & 88/103 & 85.44 \\
Object localiz. & 99/105 & 94.29 & Image emotion & 73/87 & 83.91 \\
Action recog. & 110/117 & 94.02 & Social relation & 76/91 & 83.52 \\
Identity reason. & 71/76 & 93.42 & Future predict. & 59/74 & 79.73 \\
Celebrity recog. & 135/147 & 91.84 & Image quality & 90/117 & 76.92 \\
Image scene & 131/144 & 90.97 & Physical rel. & 57/75 & 76.00 \\
Attribute recog. & 78/88 & 88.64 & Phys. prop. & 54/73 & 73.97 \\
\bottomrule
\end{tabular*}
\par\smallskip
\begin{minipage}{\linewidth}\footnotesize
Category counts ($n/N$) give correct groups out of all groups. CircularEval scores groups. Vanilla scores individual rows. Both DEV runs agreed on group correctness for all 590 training-disjoint groups, including the 326 outside all v3 splits. With images withheld, the training-disjoint DEV score was 34.58\% (204/590). LiquidAI reports 86.50\% for Qwen3.5-4B on full DEV.
\end{minipage}
\end{table}

\FloatBarrier
\section{Complete Group-Level Paired Statistical Significance Tests}
\label{app:sig}

Table~\ref{tab:app_sig} reports 15 paired group-level comparisons across official benchmarks, controlled ablations, closed-loop actions, and cross-domain transfer.

\begin{table}[tbp]
\centering\tableformat
\setlength{\tabcolsep}{2pt}
\caption{Group-level paired comparisons. Each change ($\Delta$) is the BiArbiter-4B score minus its comparator. Confidence intervals use 10,000 paired bootstrap resamples (2,000 for RewardBench v1 overall and RuVerBench AC).}
\label{tab:app_sig}
\begin{tabular*}{\linewidth}{@{\extracolsep{\fill}}lrrrrrr@{}}
\toprule
Comparison & $N$ & From & To & $\Delta$ & 95\% CI & $p$ \\
\midrule
\multicolumn{7}{@{}l}{\textit{Benchmarks and ablations}} \\
RB v1 overall & 2,985 & 86.3 & 89.1 & +2.81 & [+1.68,+3.92] & 0.0005\textsuperscript{b} \\
RB v1 raw & 2,985 & 87.2 & 90.0 & +2.78 & [+1.88,+3.72] & $<10^{-8}$\textsuperscript{m} \\
RB 2 vs. A0 & 1,812 & 66.7 & 71.9 & +5.19 & [+2.85,+7.51] & 0.0001\textsuperscript{c} \\
RB 2 vs. base & 1,812 & 72.6 & 71.9 & $-0.70$ & [$-2.64$,+1.22] & 0.4866\textsuperscript{c} \\
RuVerBench AC & 210 & 58.0 & 65.8 & +7.80 & [+1.39,+13.5] & 0.0180\textsuperscript{c} \\
Routing F1 & 22 & 0.027 & 0.858 & +0.832 & [+0.75,+0.90] & 0.0001\textsuperscript{p} \\
MMBench vs. blind & 590 & 34.6 & 90.5 & +55.9 & [+51.7,+60.3] & $<10^{-10}$\textsuperscript{m} \\
RB v1 vs. A4 & 2,985 & 89.6 & 90.0 & +0.37 & [+0.13,+0.60] & 0.0034\textsuperscript{m} \\
\addlinespace
\multicolumn{7}{@{}l}{\textit{SWE-agent issue resolution}} \\
Mixed vs. Pass@1 & 94 & 43.6 & 75.5 & +31.9 & [+23.5,+39.9] & 0.0001\textsuperscript{p} \\
Mixed vs. base & 94 & 69.2 & 75.5 & +6.38 & [$-2.13$,+14.9] & 0.2101\textsuperscript{m} \\
Mixed, $\tau=0$ & 94 & 52.1 & 66.0 & +13.8 & [+4.25,+24.5] & 0.0146\textsuperscript{m} \\
Natural vs. Pass@1 & 200 & 9.5 & 15.0 & +5.50 & [+2.75,+8.44] & 0.0001\textsuperscript{p} \\
\addlinespace
\multicolumn{7}{@{}l}{\textit{Held-out retrieval and ranking}} \\
MRAG all & 1,353 & 32.6 & 36.4 & +3.84 & [+1.55,+6.14] & 0.0013\textsuperscript{m} \\
MRAG mixed & 731 & 56.9 & 64.0 & +7.11 & [+2.87,+11.49] & 0.0013\textsuperscript{m} \\
RubricBench & 1,146 & 44.9 & 49.0 & +4.19 & [+0.09,+8.20] & 0.0485\textsuperscript{m} \\
\bottomrule
\end{tabular*}
\par\smallskip
\begin{minipage}{\linewidth}\footnotesize
Changes are percentage points except for routing F1 (0--1). The comparator is the frozen base unless specified (original prompt for Routing F1). RewardBench~2 clusters 1,865 prompts into 1,812 independent resampling units. MMBench uses the blind-image model, MRAG uses CLIP, and RubricBench uses the length heuristic. SWE-agent uses best-of-eight except in the zero-threshold early-stopping comparison. Superscripts denote bootstrap (b), cluster or case bootstrap (c), exact two-sided McNemar (m), and paired sign-flip permutation (p).
\end{minipage}
\end{table}

\FloatBarrier
\section{Cross-Domain Sub-Category Details and Zero-Overlap Audit}
\label{app:crossdomain}

Table~\ref{tab:app_cross} details per-aspect MRAG-Bench retrieval/QA behavior, per-category MVTec-AD defect detection, official five-group RubricBench accuracy, and the verified SHA-256 zero-overlap boundaries enforced across all training stages.

\begin{table}[tbp]
\centering\tableformat
\setlength{\tabcolsep}{2pt}
\caption{Subgroup performance on three held-out benchmarks and overlap checks for transfer evaluation.}
\label{tab:app_cross}
\begin{tabular*}{\linewidth}{@{\extracolsep{\fill}}lrlrrr@{}}
\toprule
Subgroup & $n$ & Metric & Simple & Base & BiArbiter \\
\midrule
\multicolumn{6}{@{}l}{\textit{MRAG-Bench}} \\
Perspective & 778 & Top-1 hit & 39.2 & 45.5 & 43.4 \\
 &  & QA accuracy & 56.7 & 54.0 & 53.7 \\
Transformative & 455 & Top-1 hit & 16.3 & 22.0 & 20.7 \\
 &  & QA accuracy & 46.4 & 50.5 & 49.5 \\
Other & 120 & Top-1 hit & 51.7 & 54.2 & 50.8 \\
 &  & QA accuracy & 51.7 & 48.3 & 49.2 \\
\addlinespace
\multicolumn{6}{@{}l}{\textit{MVTec-AD}} \\
Bottle & 83 & AUROC & 0.500 & 0.890 & 0.884 \\
 &  & BAcc & 50.0 & 77.0 & 76.2 \\
Metal nut & 115 & AUROC & 0.500 & 0.852 & 0.851 \\
 &  & BAcc & 50.0 & 55.4 & 53.8 \\
\addlinespace
\multicolumn{6}{@{}l}{\textit{RubricBench (no rubric supplied)}} \\
Chat & 422 & Pair accuracy & -- & 52.37 & 49.53 \\
Code & 270 & Pair accuracy & -- & 55.56 & 53.70 \\
STEM & 250 & Pair accuracy & -- & 53.20 & 45.20 \\
IF & 124 & Pair accuracy & -- & 52.42 & 51.61 \\
Safety & 80 & Pair accuracy & -- & 35.00 & 38.75 \\
\bottomrule
\end{tabular*}
\par\smallskip
\begin{minipage}{\linewidth}\footnotesize
Accuracy and retrieval hit rates are percentages. AUROC ranges from 0 to 1. Simple denotes original CLIP top-1 retrieval (or QA using original CLIP top-1) for MRAG-Bench, and random chance (0.500 AUROC, 50.0\% BAcc) for MVTec-AD. MRAG top-1 gains over CLIP were 4.24 points for Perspective and 4.40 for Transformative. Transformative QA improved by 3.08 points. RubricBench safety accuracy rose by 3.75 points over base.

\smallskip
\textit{Secondary transfer metrics (complementing Table~\ref{tab:transfer}).} MRAG-Bench all ($n=1{,}353$) MRR: Simple 0.414, Base 0.455, BiArbiter-4B 0.443. MRAG-Bench mixed ($n=731$) NDCG@5: Simple 0.789, Base 0.850 (direct rounding of 0.85047), BiArbiter-4B 0.833. MP-DocVQA ($n=3{,}554$) MRR: Base 0.846, BiArbiter-4B 0.839. MVTec-AD ($n=198$) macro BAcc: Simple 50.0\%, Base 66.2\%, BiArbiter-4B 65.0\%. IFBench ($n=295$) BAcc: strict Base 64.6\% vs.\ BiArbiter-4B 65.0\%; loose Base 66.6\% vs.\ BiArbiter-4B 66.2\%.

\smallskip
\textit{Overlap checks.} Stage 4 training had zero matches against RewardBench prompts, RuVerBench prompts and rows, v3 development and holdout hashes, and SWE-agent shard 04--05 issue IDs. MVTec-AD had zero matches against 688 v3 training images. Filtering excluded 1 of 1,147 RubricBench items and 5 of 300 IFBench items.
\end{minipage}
\end{table}

\FloatBarrier
\section{Routing and Verification Diagnostics by Model Variant}
\label{app:evolution_diagnostics}

\begin{table}[tbp]
\centering\tableformat
\setlength{\tabcolsep}{2pt}
\caption{Routing and verification diagnostics for the seven model variants in Table~\ref{tab:main_evolution}.}
\label{tab:evolution_diagnostics}
\begin{tabular*}{\linewidth}{@{\extracolsep{\fill}}lrrr@{}}
\toprule
Model variant & \makecell[r]{Empty dispatches\\(of 80)} & \makecell[r]{DeepResearch\\AUROC} & \makecell[r]{AgenticCoding\\AUROC} \\
\midrule
A0, Causal base & 47 & 0.9315 & 0.7520 \\
A1, Bidirectional base & 42 & 0.9392 & 0.7437 \\
A2, Causal LoRA & 17 & 0.9275 & 0.7360 \\
A3, Bidirectional LoRA & 11 & 0.9370 & 0.7337 \\
Step 6, Balanced roster & 2 & 0.9407 & 0.7341 \\
Step 7, Hard negatives & 0 & 0.9432 & 0.7279 \\
BiArbiter-4B & 0 & 0.9442 & 0.7106 \\
\bottomrule
\end{tabular*}
\par\smallskip
\begin{minipage}{\linewidth}\footnotesize
Empty dispatches count InFoBench requirements for which the model selected no checker. Both AUROCs are from RuVerBench.
\end{minipage}
\end{table}
Table~\ref{tab:main_evolution} uses the refined checker roster for A0--A3. With the original prompt, their routing macro-F1 scores were 0.036, 0.027, 0.220, and 0.374, respectively. Independent reruns and split audits are reported in Tables~\ref{tab:app_ruver} and~\ref{tab:app_mmbench}.

\FloatBarrier
\section{Complete Tabular Records Supporting Figures~\ref{fig:ablations}--\ref{fig:downstream}}
\label{app:migrated_tables}

Tables~\ref{tab:supp_ablation}--\ref{tab:supp_downstream} preserve the full numerical values visualized in Figures~\ref{fig:ablations}--\ref{fig:downstream}, enabling exact cell-level lookup complementing the visualizations in the main text.

\begin{table}[tbp]
\centering\tableformat
\caption{Stage-4 training ablations from the same Stage~3 midpoint (Step~7) checkpoint (supporting Figure~\ref{fig:ablations}). Each trained variant uses the same 2,781 training sets, 348 updates, and random seed.}
\label{tab:supp_ablation}
\setlength{\tabcolsep}{2pt}
\begin{tabular*}{\linewidth}{@{\extracolsep{\fill}}lrrrr@{}}
\toprule
& & \multicolumn{2}{c}{Matched removals} & \\
\cmidrule(lr){3-4}
Metric & Step 7 & \makecell[r]{No anchor\\(A4)} & \makecell[r]{Symmetric\\gate (A5)} & \textbf{Full} \\
\midrule
\multicolumn{5}{@{}l}{\textit{Parameter drift and preference ranking}} \\
Squared LoRA drift & 0.0000 & 0.2267 & 0.1446 & 0.1574 \\
RewardBench v1 overall (\%) & 88.50 & 88.80 & 89.01 & 89.09 \\
Raw pair accuracy (\%) & 89.41 & 89.58 & 89.88 & 89.95 \\
\midrule
\multicolumn{5}{@{}l}{\textit{RuVerBench verification}} \\
DeepResearch BAcc (\%) & 85.59 & 85.78 & 85.96 & 85.84 \\
DeepResearch AUROC & 0.9432 & 0.9441 & 0.9435 & 0.9442 \\
AgenticCoding BAcc (\%) & 62.72 & 65.67 & 65.58 & 65.77 \\
Planning BAcc (\%) & 63.5 & 71.8 & 70.8 & 73.1 \\
Predicted positive (\%) & 84.2 & 71.2 & 74.6 & 71.8 \\
Negative recall, micro (\%) & 33.1 & 49.7 & 45.5 & 48.3 \\
\midrule
\multicolumn{5}{@{}l}{\textit{Visual choice, routing, and agent verification}} \\
MMBench DEV, unseen (\%) & 90.51 & 90.51 & 90.34 & 90.51 \\
Routing macro-F1 & 0.8517 & 0.8533 & 0.8634 & 0.8584 \\
SWE dev BAcc (\%) & 65.0 & 68.0 & 67.0 & 69.0 \\
SWE dev negative recall (\%) & 38.0 & 74.0 & 68.0 & 74.0 \\
\bottomrule
\end{tabular*}
\par\smallskip
\begin{minipage}{\linewidth}\footnotesize
Full denotes BiArbiter-4B. Squared LoRA drift is computed directly from final saved adapter weights (0.157415; pre-final-step training log recorded 0.1573, both giving a 30.6\% reduction relative to A4 at 0.226732). A4 removes the $L_2$ anchor; A5 removes asymmetric negative weighting and pairwise gate loss. AgenticCoding negative recall is pooled (micro) over all 145 negative checks; four-category macro negative recall rises from 36.04\% (Step 7) to 55.37\% (Full).

\smallskip
\textit{Image ablation at inference.} On the 590 training-disjoint MMBench DEV groups, Full scored 90.51\% with images (534/590) and 34.58\% without (204/590), yielding $+344/-14$ paired group changes ($+55.93$\,pp, McNemar $p<10^{-10}$).
\end{minipage}
\end{table}

\begin{table}[tbp]
\centering\tableformat
\setlength{\tabcolsep}{1.8pt}
\caption{Measured inference cost of BiArbiter-4B and causal CoT judging on the same 4B backbone on one NVIDIA A100-SXM4-80GB MIG \texttt{3g.40gb} slice in BF16/SDPA (supporting Figure~\ref{fig:efficiency}).}
\label{tab:supp_efficiency}
\begin{tabular*}{\linewidth}{@{\extracolsep{\fill}}lrrr@{}}
\toprule
\multicolumn{4}{@{}l}{\textit{Single-candidate latency and peak GPU memory}} \\
\midrule
Scoring condition & p50 (ms) & p95 (ms) & Peak (GiB) \\
\midrule
One pass, 256 tokens & 61.90 & 62.74 & 8.62 \\
One pass, 373 tokens (image + text) & 89.00 & 91.09 & 8.65 \\
One pass, 1,024 tokens & 132.57 & 132.81 & 8.72 \\
One pass, 4,096 tokens & 529.11 & 529.43 & 9.13 \\
One pass, 8,192 tokens & 1,100.84 & 1,102.02 & 9.67 \\
\addlinespace
CoT, 128 output tokens & 5,352.63 & 5,510.27 & 8.73 \\
CoT, 256 output tokens & 10,622.43 & 10,701.52 & 8.73 \\
\bottomrule
\end{tabular*}
\par\medskip
\begin{tabular*}{\linewidth}{@{\extracolsep{\fill}}lrrrrr@{}}
\toprule
 & \multicolumn{3}{c}{Batched scoring} & \multicolumn{2}{c}{Single-candidate speedup} \\
\cmidrule(lr){2-4}\cmidrule(l){5-6}
Input tokens & Batch & Cand./s & Peak (GiB) & CoT 128 & CoT 256 \\
\midrule
256 & 8 & 31.11 & 8.98 & 86.5$\times$ & 171.6$\times$ \\
373 (image + text) & -- & -- & -- & 60.1$\times$ & 119.4$\times$ \\
1,024 & 8 & 8.03 & 9.67 & \textbf{40.4}$\times$ & \textbf{80.1}$\times$ \\
4,096 & 4 & 1.93 & 10.75 & 10.1$\times$ & 20.1$\times$ \\
8,192 & 4 & 0.92 & 12.91 & 4.9$\times$ & 9.6$\times$ \\
\bottomrule
\end{tabular*}
\par\smallskip
\begin{minipage}{\linewidth}\footnotesize
Both CoT runs used a 789-token prompt with KV-cached greedy decoding. Speedup equals CoT p50 latency divided by one-pass p50 latency at batch size 1. The 373-token input included a synthetic 512-pixel RGB image. Dashes indicate unmeasured batch configurations.
\end{minipage}
\end{table}

\begin{table}[tbp]
\centering\tableformat
\setlength{\tabcolsep}{2pt}
\caption{Calibration and selective prediction for the frozen bidirectional base (A1) and BiArbiter-4B (supporting Figure~\ref{fig:calibration}).}
\label{tab:supp_calibration}
\begin{tabular*}{\linewidth}{@{\extracolsep{\fill}}lrrrrr@{}}
\toprule
Metric & Route & DR & RB v1 & MM DEV & AC \\
\midrule
\multicolumn{6}{@{}l}{\textit{Frozen base (A1)}} \\
Raw ECE & 0.1156 & 0.0794 & 0.0777 & 0.0411 & 0.1099 \\
Dev-scaled ECE & 0.0293 & 0.1271 & 0.1435 & -- & 0.1509 \\
Brier score & 0.1817 & 0.1036 & 0.1076 & 0.0719 & 0.1562 \\
Accuracy (\%) & 76.75 & 85.45 & 87.17 & 90.40 & 78.65 \\
AURC & 0.1548 & 0.0445 & 0.0450 & 0.0173 & 0.1168 \\
Accuracy at 80\% coverage (\%) & 80.62 & 92.26 & 92.17 & 96.72 & 83.23 \\
Accuracy at 70\% coverage (\%) & 81.07 & 94.07 & 93.59 & 97.21 & 84.75 \\
\midrule
\multicolumn{6}{@{}l}{\textit{BiArbiter-4B}} \\
Raw ECE & 0.0428 & 0.0429 & 0.0691 & 0.0297 & 0.1526 \\
Dev-scaled ECE & 0.0683 & 0.0843 & 0.1289 & -- & 0.1714 \\
Brier score & 0.0591 & 0.0943 & 0.0910 & 0.0641 & 0.1905 \\
Accuracy (\%) & 92.50 & 86.63 & 89.95 & 90.89 & 71.17 \\
AURC & 0.0105 & 0.0388 & 0.0365 & 0.0150 & 0.1485 \\
Accuracy at 80\% coverage (\%) & 97.81 & 93.27 & 93.68 & 97.11 & 77.89 \\
Accuracy at 70\% coverage (\%) & 99.64 & 94.87 & 95.02 & 97.84 & 80.68 \\
\bottomrule
\end{tabular*}
\par\smallskip
\begin{minipage}{\linewidth}\footnotesize
Route covers 80 InFoBench requirements (400 binary checks), DR covers 1,615 DeepResearch rubrics, RB v1 covers 2,985 RewardBench pairs, MM DEV covers 590 training-disjoint groups (2,250 circular-eval rows), and AC covers 843 AgenticCoding checks (82.8\% positive, 698/843). ECE uses 15 equal-width bins; Dev-scaled ECE applies routing-dev temperature $T_{\mathrm{dev}}=1.6042$. Lower ECE, Brier, and AURC are better. Across all four evaluated checkpoints (frozen \texttt{base}, Stage~2 / \texttt{step6}, Stage~3 midpoint / \texttt{step7\_mid}, and Stage~4 / \textbf{BiArbiter-4B}), BiArbiter-4B achieves the strictly lowest AURC on all four non-imbalanced splits; intermediate checkpoints achieve slightly lower raw ECE on RB v1 (\texttt{step6} 0.0580) and MM DEV (\texttt{step7\_mid} 0.0279) and slightly lower Brier on Route (\texttt{step6} 0.0579) and MM DEV (\texttt{step7\_mid} 0.0639).
\end{minipage}
\end{table}

\begin{table}[tbp]
\centering\tableformat
\setlength{\tabcolsep}{2pt}
\caption{Checker routing on InFoBench (80 requirements across 22 tasks) and unit-test resolution on held-out SWE-agent shards 04--05 ($K=8$, zero shared issue IDs with train/dev; supporting Figure~\ref{fig:downstream}).}
\label{tab:supp_downstream}
\begin{tabular*}{\linewidth}{@{\extracolsep{\fill}}lrrrrrr@{}}
\toprule
Routing policy & Macro-F1 & \makecell[r]{Exact\\$n$ (\%)} & Empty & \makecell[r]{Conflicts\\of 8} & Calls & \makecell[r]{Saved\\(\%)} \\
\midrule
All five & 0.3673 & 0 (0.0) & 0 & 0 & 5.00 & 0.0 \\
Content only & 0.6185 & 37 (46.2) & 0 & 0 & 1.00 & 80.0 \\
Random & 0.1898 & -- & 19.2 & 1.5 & 1.24 & 75.2 \\
Base (original) & 0.0269 & 3 (3.8) & 75 & 0 & 0.10 & 98.0 \\
Step 6 & 0.8394 & 56 (70.0) & 2 & 5 & 1.25 & 75.0 \\
BiArbiter-4B & 0.8584 & 56 (70.0) & 0 & 7 & 1.24 & 75.2 \\
\bottomrule
\end{tabular*}
\par\medskip
\begin{tabular*}{\linewidth}{@{\extracolsep{\fill}}lrrr@{}}
\toprule
SWE-agent outcome (\%) & Base & Step 7 & BiArbiter-4B \\
\midrule
\multicolumn{4}{@{}l}{\textit{Natural pool (200 issues)}} \\
Best-of-8 resolution & 13.50 & 15.00 & 15.00 \\
Resolution at $q_{50}$ & 14.00 & 14.50 & 15.00 \\
Rollouts saved at $q_{50}$ & 21.9 & 20.4 & 22.1 \\
\addlinespace
\multicolumn{4}{@{}l}{\textit{Mixed pool (94 issues)}} \\
Best-of-8 resolution & 69.15 & 72.34 & 75.53 \\
Resolution at $\tau=0$ & 52.13 & 48.94 & 65.96 \\
Rollouts saved at $\tau=0$ & 80.3 & 82.4 & 56.0 \\
Resolution at $q_{50}$ & 69.15 & 70.21 & 74.47 \\
Rollouts saved at $q_{50}$ & 42.4 & 42.2 & 48.4 \\
\bottomrule
\end{tabular*}
\par\smallskip
\begin{minipage}{\linewidth}\footnotesize
Empty counts requirements with no dispatch; Calls are per requirement; Random averages 1,000 matched-budget draws. Natural and Mixed SWE pools share 34 issue IDs and are reported separately (both have zero overlap with train/dev). Expected Pass@1 resolves 9.50\% / 43.62\%, longest-trajectory resolves 9.00\% / 29.79\%, and oracle Pass@8 reaches 19.00\% / 100.00\%. Early stopping uses $\tau=0$ or each model's development-set positive median ($q_{50}$: Base 1.8051, Step 7 2.1716, BiArbiter-4B 0.2588). Step 7 Mixed savings use direct integer rounding ($620/752=82.45\%\to 82.4\%$; $317/752=42.15\%\to 42.2\%$; two-decimal archived-report values round to 82.5\% and 42.1\%).
\end{minipage}
\end{table}

\FloatBarrier
\section{Readout Controls and MRAG-Bench Test Item 2 Replay Trace}
\label{app:figure_diagnostics}

\begin{figure}[tbp]
    \centering
    \includegraphics[width=0.98\linewidth]{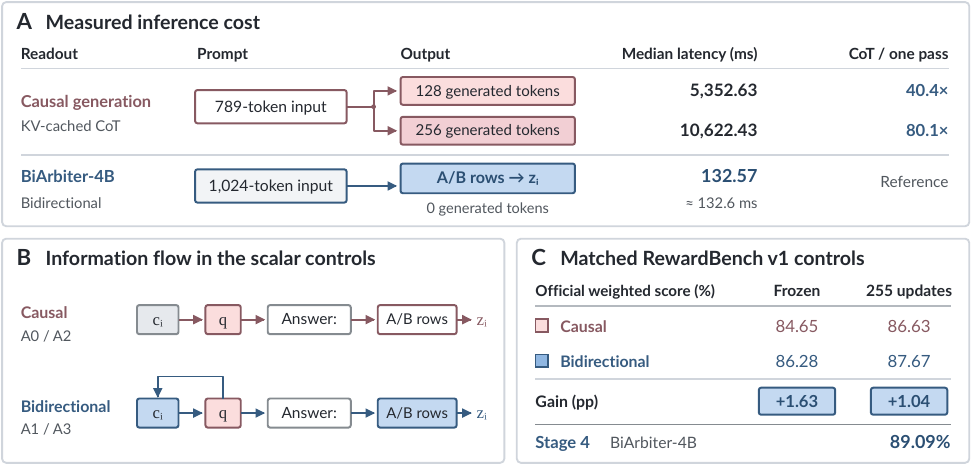}
    \caption{Readout controls on H2O-Lightning-4B. (A) Median single-candidate latency (A100 MIG \texttt{3g.40gb}, BF16, batch~1) for causal generation with a 789-token input and fixed output lengths, and for BiArbiter-4B scoring a 1,024-token input. (B) Arrows show information flow: both scalar readouts use the criterion at the last position, whereas bidirectional full attention also gives earlier candidate states access to it. (C) Differences are bidirectional minus causal RewardBench v1 scores in percentage points, first for the frozen pair and then after 255 LoRA updates per arm; Stage 4 reports the final bidirectional model separately.}
    \label{fig:comparison}
\end{figure}

Figure~\ref{fig:comparison} compares causal generation versus one-pass bidirectional scoring controls and matched RewardBench v1 attention ablations from Table~\ref{tab:main_evolution}.

\FloatBarrier
Figure~\ref{fig:case} (and Figure~\ref{fig:mrag}(C)) uses the smallest ID among 15 questions with both a corrected evidence hit and a corrected answer in the 1,353-question archive.
Its photographs and city options come from the \emph{Temporal} scenario of MRAG-Bench~\cite{hu2025mragbench}.
The single-item replay used the frozen \texttt{final-step-1035} checkpoint on an A100 MIG \texttt{2g.20gb}, whereas the full archive used a \texttt{3g.40gb} slice.
The largest absolute difference between their five evidence logits was 0.05128; both runs selected R4 and gave identical three-condition correctness outcomes.
Table~\ref{tab:case_trace} reports all 17 replay scores.

\begin{table}[tbp]
\centering\tableformat
\caption{MRAG-Bench test ID 2. Five evidence scores and twelve city scores in the single-item replay (supporting Figure~\ref{fig:case}). Bold marks the largest value in each decision.}
\label{tab:case_trace}
\begin{tabular*}{\linewidth}{@{\extracolsep{\fill}}lrrrrr@{}}
\toprule
Evidence & R1 & R2 & R3 & R4 & R5 \\
\midrule
Logit & $-0.22620$ & $-1.18142$ & $-0.85688$ & $\mathbf{-0.16401}$ & $-0.20108$ \\
\bottomrule
\end{tabular*}
\par\medskip
\begin{tabular*}{\linewidth}{@{\extracolsep{\fill}}lrrr@{}}
\toprule
City candidate & Query only & Query + R1 & Query + R4 \\
\midrule
Boston        & $0.21787$ & $-0.70887$ & $-0.83396$ \\
Chicago       & $\mathbf{0.83537}$ & $-0.84834$ & $0.81669$ \\
New York City & $0.65617$ & $-1.27914$ & $\mathbf{1.29758}$ \\
Philadelphia  & $0.65911$ & $\mathbf{-0.27482}$ & $0.01622$ \\
\bottomrule
\end{tabular*}
\par\smallskip
\begin{minipage}{\linewidth}\footnotesize
For R4, the two vocabulary logits are 5.86494 and 5.98795; their difference divided by 0.75 gives the displayed evidence score to replay precision. The reader selects the city with the highest score in its column. The benchmark answer is New York City.
\end{minipage}
\end{table}

\end{document}